\documentclass[journal,hideappendix]{vgtc}        

\onlineid{1356}

\vgtccategory{Research}

\vgtcpapertype{algorithm}

\title{Gesture2Text: A Generalizable Decoder for Word-Gesture Keyboards in XR Through Trajectory Coarse Discretization and Pre-training}

\author{%
  Junxiao Shen, Khadija Khaldi, Enmin Zhou, Hemant Bhaskar Surale, and Amy Karlson
}

\authorfooter{
  \item
  	Junxiao Shen is with Reality Labs Research, Meta and University of Bristol.
  \item
        Khadija Khaldi, Enmin Zhou, Hemant Bhaskar Surale and Amy Karlson are with Reality Labs Research, Meta.

}

\abstract{Text entry with word-gesture keyboards (WGK) is emerging as a popular method and becoming a key interaction for Extended Reality (XR). However, the diversity of interaction modes, keyboard sizes, and visual feedback in these environments introduces divergent word-gesture trajectory data patterns, thus leading to complexity in decoding trajectories into text. Template-matching decoding methods, such as SHARK$^2$~\cite{kristensson2004shark2}, are commonly used for these WGK systems because they are easy to implement and configure. However, these methods are susceptible to decoding inaccuracies for noisy trajectories. While conventional neural-network-based decoders (neural decoders) trained on word-gesture trajectory data have been proposed to improve accuracy, they have their own limitations: they require extensive data for training and deep-learning expertise for implementation. To address these challenges, we propose a novel solution that combines ease of implementation with high decoding accuracy: a generalizable neural decoder enabled by pre-training on large-scale coarsely discretized word-gesture trajectories. This approach produces a ready-to-use WGK decoder that is generalizable across mid-air and on-surface WGK systems in augmented reality (AR) and virtual reality (VR), which is evident by a robust average Top-4 accuracy of 90.4\% on four diverse datasets. It significantly outperforms SHARK$^2$ with a 37.2\% enhancement and surpasses the conventional neural decoder by 7.4\%. Moreover, the \textit{Pre-trained Neural Decoder}'s size is only 4 MB after quantization, without sacrificing accuracy, and it can operate in real-time, executing in just 97 milliseconds on Quest 3.
}


\teaser{
  \centering
    \begin{subfigure}[b]{0.19\textwidth}
        \includegraphics[width=\textwidth]{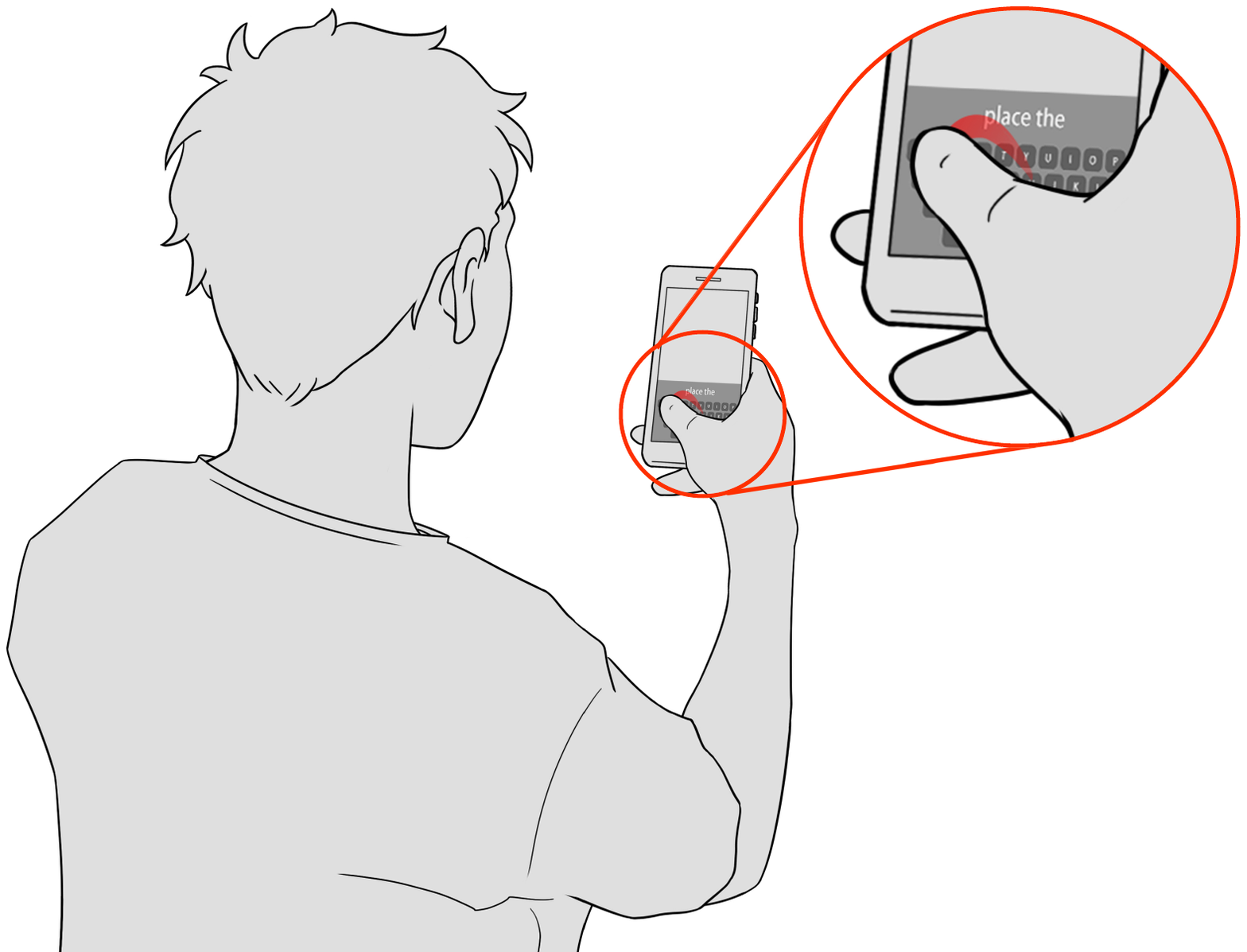}
    \end{subfigure}
    \hfill
    \begin{subfigure}[b]{0.19\textwidth}
        \includegraphics[width=\textwidth]{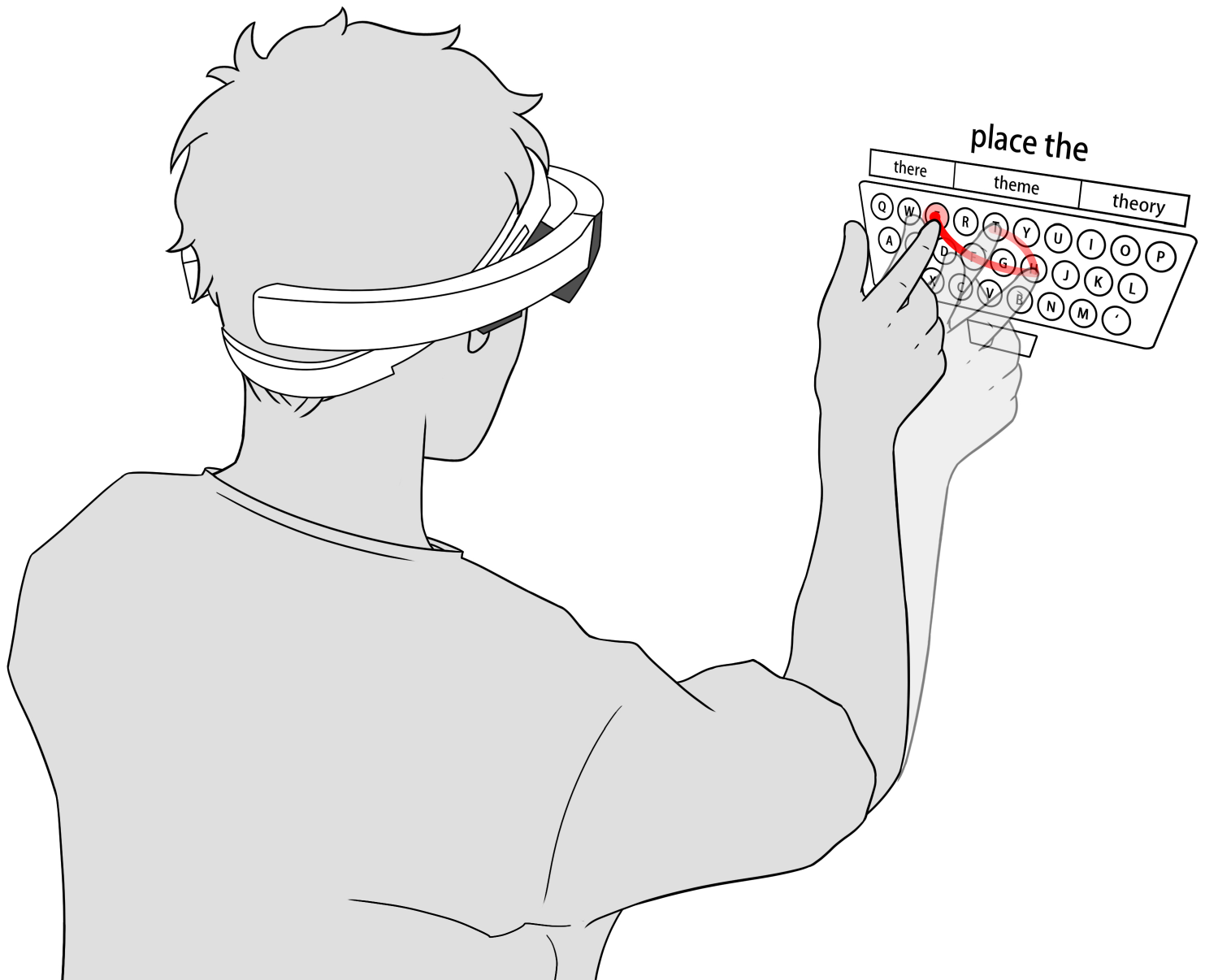}
    \end{subfigure}
    \hfill
    \begin{subfigure}[b]{0.19\textwidth}
        \includegraphics[width=\textwidth]{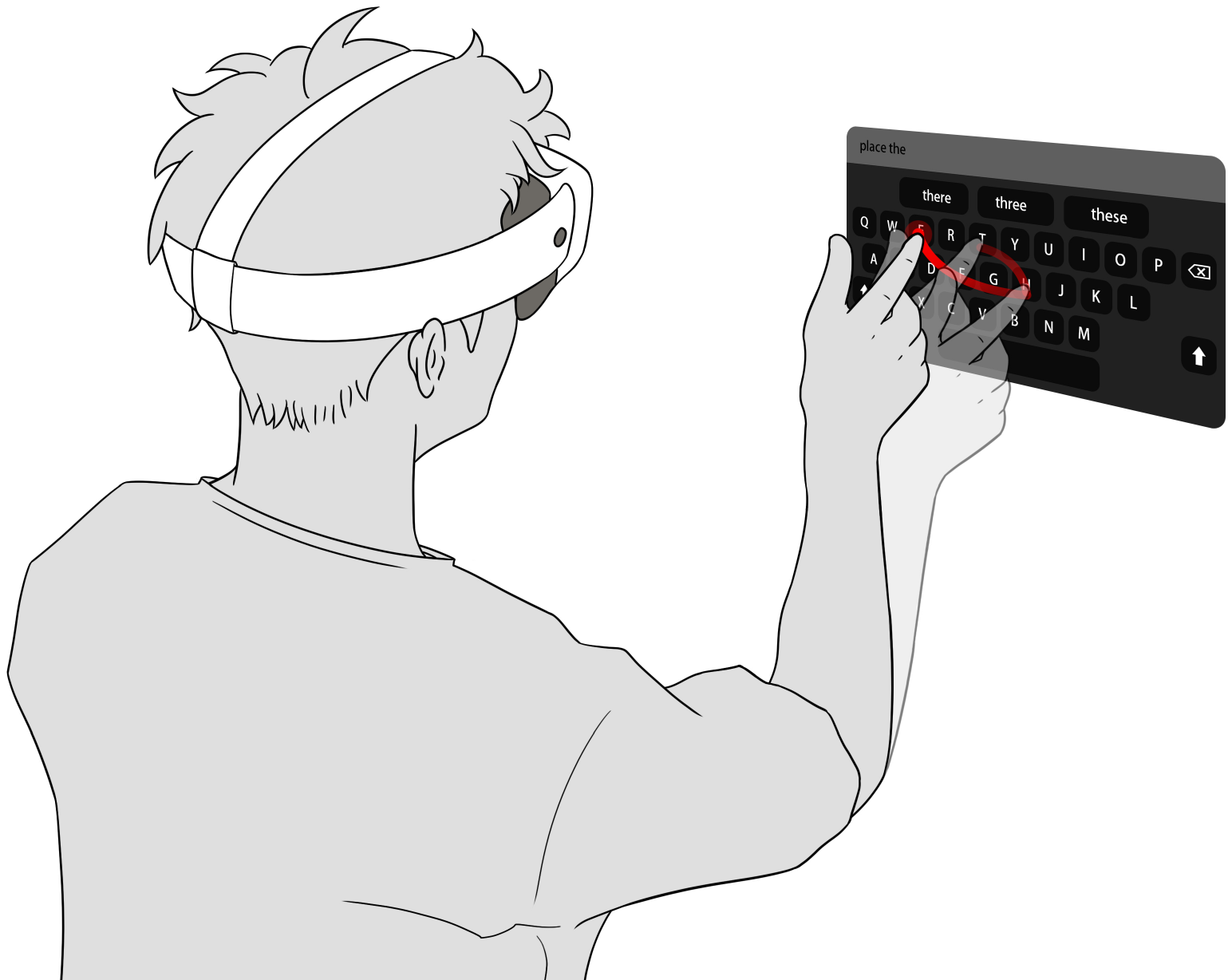}
    \end{subfigure}
    \hfill
    \begin{subfigure}[b]{0.19\textwidth}
        \includegraphics[width=\textwidth]{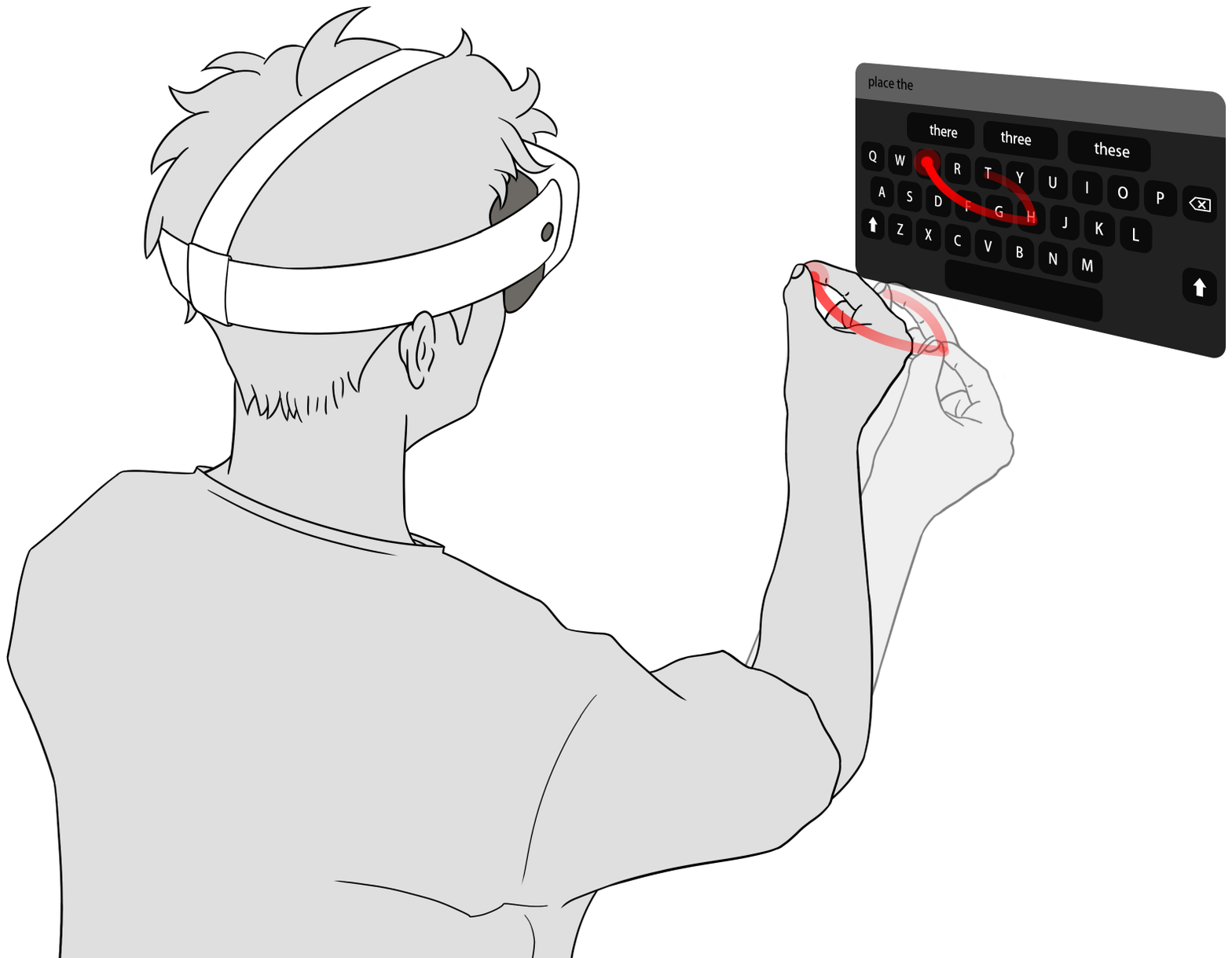}
    \end{subfigure}
    \hfill
    \begin{subfigure}[b]{0.19\textwidth}
        \includegraphics[width=\textwidth]{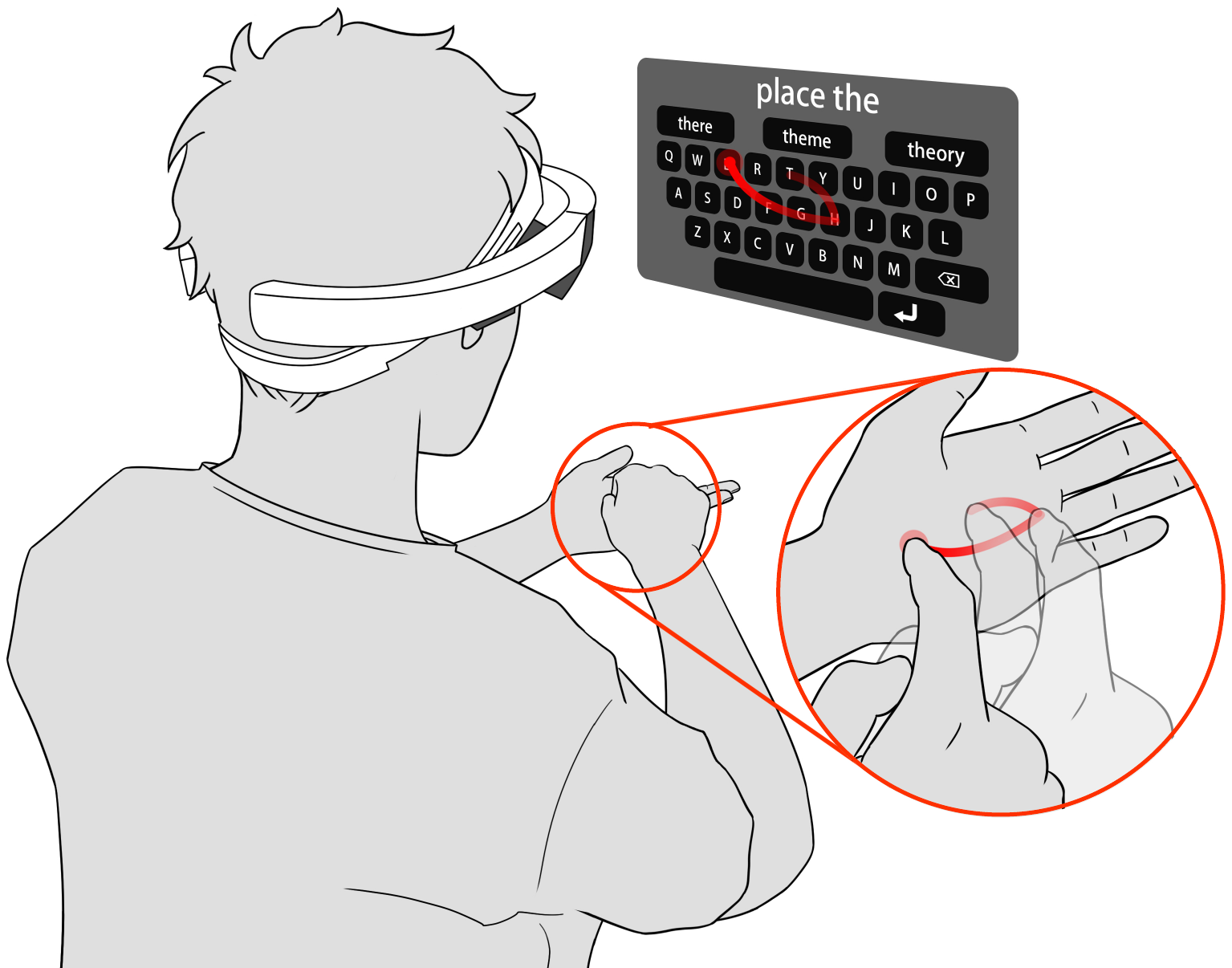}
    \end{subfigure}
    \hfill
    \begin{subfigure}[b]{0.19\textwidth}
        \includegraphics[width=\textwidth]{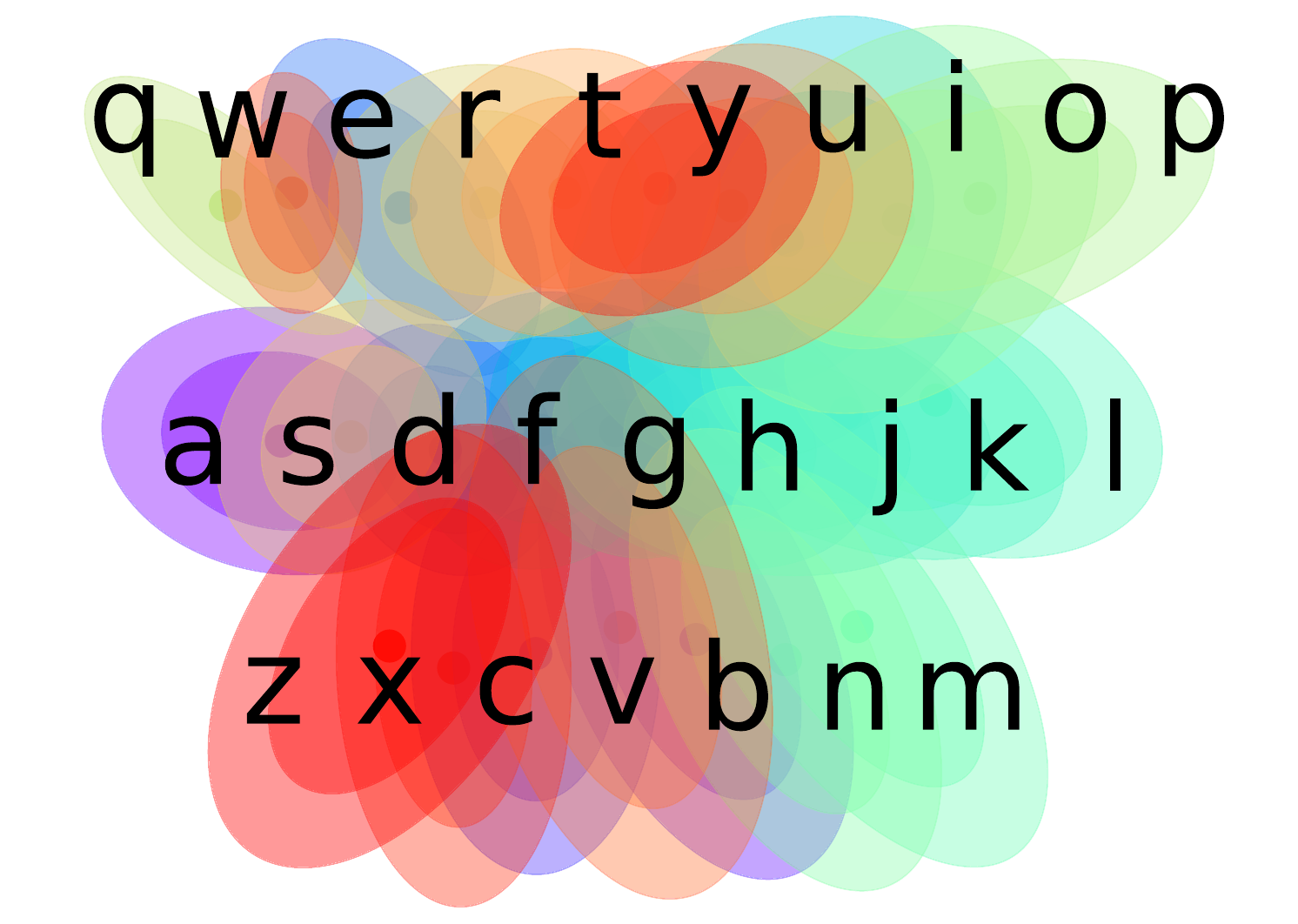}
        \caption{Mobile Phone \\ WGK}
    \end{subfigure}
    \hfill
    \begin{subfigure}[b]{0.19\textwidth}
        \includegraphics[width=\textwidth]{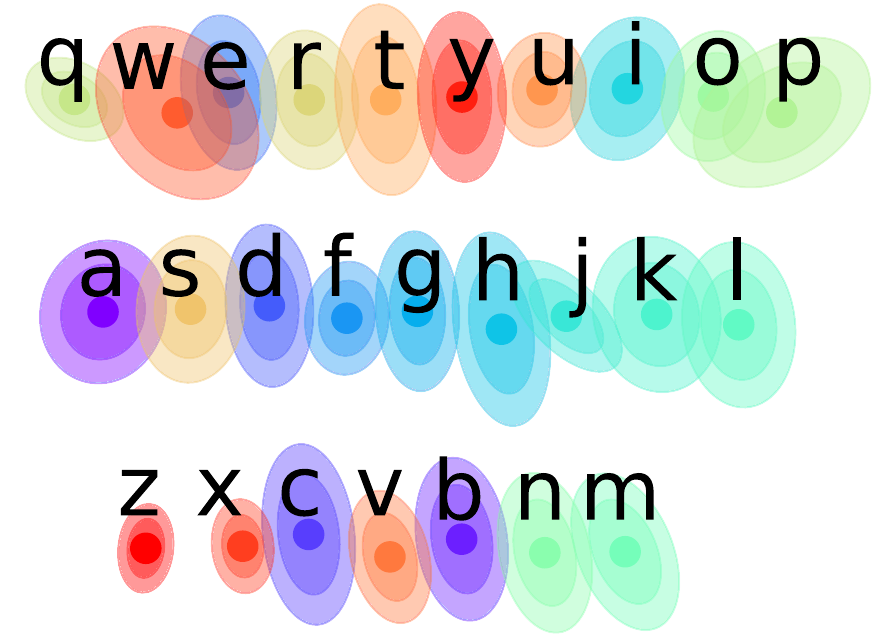}
        \caption{Mid-Air Poke \\ (AR) WGK}
    \end{subfigure}
    \hfill
    \begin{subfigure}[b]{0.19\textwidth}
        \includegraphics[width=\textwidth]{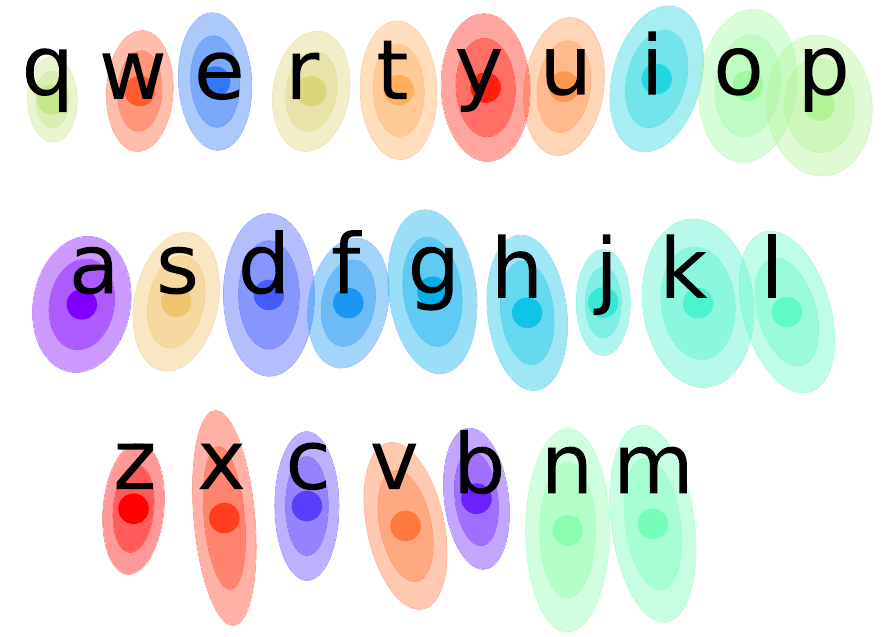}
        \caption{Mid-Air Poke \\ (VR) WGK}
    \end{subfigure}
    \hfill
    \begin{subfigure}[b]{0.19\textwidth}
        \includegraphics[width=\textwidth]{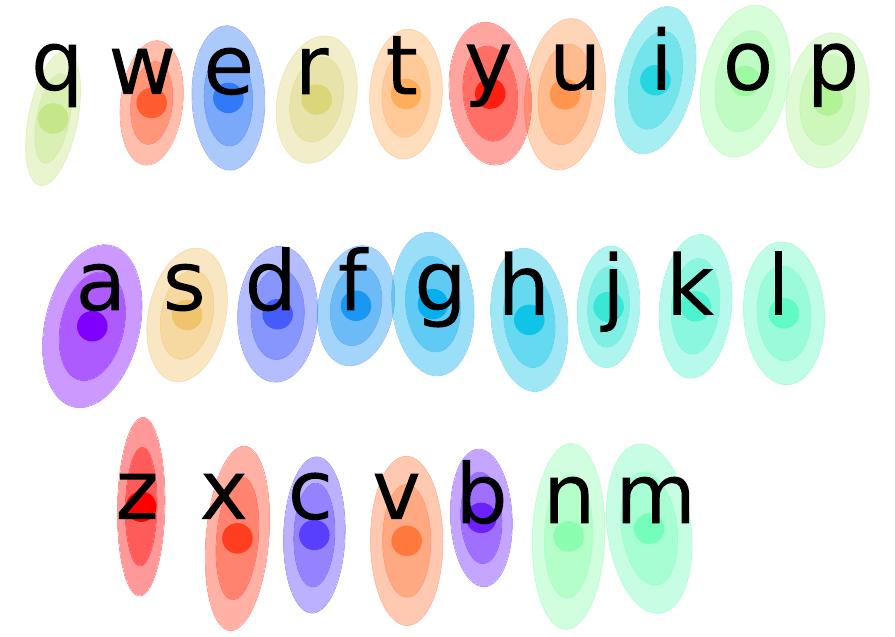}
        \caption{Mid-Air Pinch \\ (VR) WGK}
    \end{subfigure}
    \hfill
    \begin{subfigure}[b]{0.19\textwidth}
        \includegraphics[width=\textwidth]{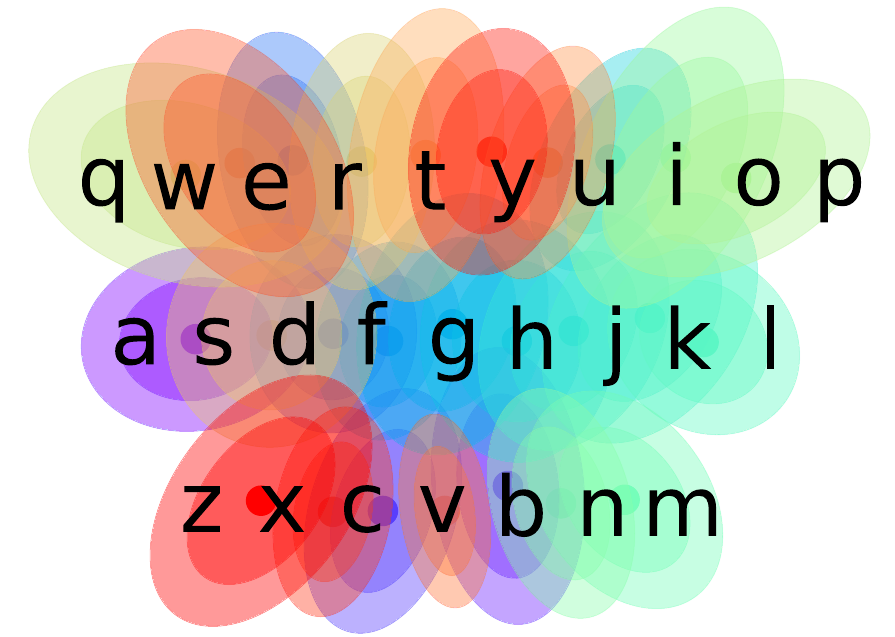}
        \caption{On-Surface \\ WGK}
    \end{subfigure}
  \caption{
We pre-trained a neural-network-based word-gesture keyboard (WGK) decoder for \textbf{Gesture2Text} decoding, also noted as a neural decoder, using a novel one-hot encoded coarse discretized trajectory representation. This approach enhances generalizability across various WGK systems illustrated above. The distribution of touch points in the second row illustrates that different WGK systems exhibit unique data patterns due to variations in interaction modes, keyboard sizes, and user behaviors. 
  }
  \label{fig:teaser}
}

\renewcommand{\manuscriptnotetxt}{}

\usepackage[svgnames]{xcolor}

\graphicspath{{figs/}{figures/}{pictures/}{images/}{./}} 

\usepackage{tabu}                      
\usepackage{booktabs}                  
\usepackage{lipsum}                    
\usepackage{mwe}                       

\usepackage{mathptmx}                  

\usepackage{microtype}                 
\PassOptionsToPackage{warn}{textcomp}  
\usepackage{textcomp}                  
\usepackage{mathptmx}                  
\usepackage{times}                     
\usepackage{cite}                      
\usepackage{tabu}                      
\usepackage{booktabs}   

\usepackage{amsmath}
\usepackage{amssymb}
\usepackage{multirow}
\usepackage{graphicx}

\usepackage{floatrow}
\usepackage{url}
\usepackage{booktabs}
\usepackage{array}
\usepackage{subcaption}
\usepackage{float}
\usepackage{dblfloatfix} 

\usepackage{amsmath}
\usepackage{algorithm}
\usepackage{algpseudocode}

\usepackage{booktabs} 
\usepackage{siunitx} 

\usepackage{tabularray}

\newenvironment{s_itemize}{
\begin{itemize}[leftmargin=*]
  \setlength{\itemsep}{3pt}
  \setlength{\parskip}{0pt}
  \setlength{\parsep}{0pt}
}{\end{itemize}}

\newenvironment{s_enumerate}{
\begin{enumerate}[wide, labelwidth=!, labelindent=0pt]
  \setlength{\itemsep}{2pt}
  \setlength{\parskip}{0pt}
  \setlength{\parsep}{0pt}
}{\end{enumerate}}

\usepackage{array}
\usepackage{enumitem}

\usepackage{enumitem}
\usepackage{todonotes}

\usepackage{subcaption}
\usepackage{float}
\usepackage{dblfloatfix}

\definecolor{NavyBlue}{RGB}{0,0,128}

\newcommand{\changetext}[1]{\textcolor{black}{#1}}

\begin{document}


\firstsection{Introduction}

\maketitle

Extensive techniques have been proposed and developed to enable fast and accurate text entry in Extended Reality (XR) environments~\cite{dudley2018fast,xu2019tiptext, peshock2014argot, gu2020qwertyring,  lee2019hibey}.
Among these, word-gesture keyboards (WGK) emerges as a promising solution, achieving text entry speeds ranging from 20 to 40 words per minute (WPM) for proficient users~\cite{dudley2023evaluating, markussen2014vulture,chen2023DRG,gupta2019rotoswype,xu2019pointing,zhao2023gaze,kern2023text,yu2017tap,shen2023fast}.
This method not only offers excellent learnability, being widely adopted on touchscreen devices~\cite{leiva2021we}, but also inherently handles noisy and ambiguous input due to the ambiguous nature of word-gesture trajectories~\cite{kristensson2004shark2}. 
This is analogous to word-gesture typing on small screens of smartwatches, which effectively addresses the `fat finger' problem~\cite{gordon2016watchwriter}.

However, the majority of previous studies focus on the development of new interaction techniques. 
The word-gesture decoding process, which translates the word-gesture trajectory into text, predominantly employs the classic SHARK$^2$~\cite{kristensson2004shark2} decoder.
SHARK$^2$~\cite{kristensson2004shark2} is a template-matching algorithm that computes the similarity of the input trajectory with the pre-defined word-gesture templates constructed from a word corpus and gives top-ranked predictions based on trajectory-template similarity. 
The popularity of template-matching decoders stems from their simplicity; they merely require pre-defining word-gesture templates and similarity metrics, enabling the matching algorithm to be used in a plug-and-play fashion. 
However, these algorithms are not without their limitations, including inability to predict out-of-vocabulary (OOV) words and lack of decoding accuracy for noisy input trajectories~\cite{reyal2015performance}.

Alsharif et al.~\cite{alsharif2015long} and Shen et al.~\cite{shen2023fast} propose training a neural-network-based decoder, or neural decoder, to decode word-gesture trajectories into text. 
Shen et al.~\cite{shen2023fast} explicitly compared neural decoders with SHARK$^2$, suggesting that neural decoders significantly outperform SHARK$^2$. 
However, the adoption of neural decoders is limited due to a significant drawback: they require substantial amounts of training data. 
We hypothesize that different WGK systems exhibit distinct trajectory patterns, rendering a neural decoder trained on one system non-generalizable to another. 
This limitation is due not only to variations in interaction modes, but also to the different keyboard sizes and user behaviors, as indicated by the touch point distributions of the on-surface and mid-air WGK systems in AR and VR in Figure~\ref{fig:teaser}. 
Moreover, it requires deep expertise to develop and train a neural decoder, as these models are more complex to build and train compared to some classic deep-learning models which already have many open-sourced codebases~\cite{wolf-etal-2020-transformers, mmdetection}.
We aim for a generalizable neural decoder that combines ease of configuration and implementation with high decoding accuracy.

To achieve this, we propose a novel trajectory representation, which is one-hot encoded coarse discretized trajectories (see example illustration in Figure~\ref{fig:pixelated_traj}).
This novel representation tolerates high noise and differences in trajectories from different WGK interaction modes, keyboard sizes, and user behaviors.
Therefore, it enables pre-training a neural decoder on datasets from one WGK system to be generalized across other WGK systems without explicit fine-tuning.
This eliminates the need for collecting training data and complicated configuration.

More specifically, we pre-trained the decoder on a public word-gesture trajectory dataset collected from Mobile Phone WGK~\cite{leiva2021we} combined with a dataset of synthetic word-gesture trajectories~\cite{shen2021simulating}.
We further validated our \textit{Pre-trained Neural Decoder} on four datasets: one publicly available dataset for mid-air WGK in AR from Shen et al.~\cite{shen2022personalization,shen2023fast} and three datasets collected ourselves, including mid-air WGK in VR with two different interaction modes and on-surface WGK (illustrated in Figure~\ref{fig:teaser}). 
Initially, we compared the \textit{Pre-trained Neural Decoder} with a conventional neural decoder from Alsharif et al.~\cite{alsharif2015long} and SHARK$^2$, demonstrating a 7.4\% improvement and a 37.2\% improvement, respectively, in Top-4 accuracy. 
To this end, the \textit{Pre-trained Neural Decoder} could be used out of the box with an average decoding Top-4 accuracy of 90.4\%, which is adequate for large-scale applications~\cite{reyal2015performance}.
We also investigated fine-tuning the pre-trained decoder, achieving a modest improvement. 
Additionally, we explored its individual components, including different discretization techniques, encoding methods and model structures.
Lastly, we conducted a model efficiency validation test by deploying the model onto Quest 3~\cite{OculusQuestSeries} through model quantization without sacrificing decoding accuracy~\cite{Han2015DeepCC}.
The latency was approximately 97 millisecond (ms), which is adequate for commercial use.

In conclusion, our contributions are as follows:
\begin{s_enumerate}
\item We propose a novel trajectory coarse discretization approach to enable the pre-training of a word-gesture neural decoder that can be generalized to on-surface and mid-air word-gesture keyboards in AR and VR.
The resulting \textit{Pre-trained Neural Decoder} is fast and easy to deploy and does not require additional data collection for training or fine-tuning.
\item We validated the \textit{Pre-trained Neural Decoder} on datasets from four different AR/VR word-gesture keyboard systems and showed that the \textit{Pre-trained Neural Decoder} performs significantly better than conventional decoders, with a Top-4 word prediction accuracy of 90.4\%.
\item 
We tested the real-time performance of the pre-trained decoder on a mobile VR device, Quest 3, by quantizing the model, suggesting low latency of 97 ms and real-world applicability of the model.
\end{s_enumerate}

\section{Related Work}
\subsection{Word-Gesture Keyboards}
The word-gesture keyboard (WGK), originally developed for text entry on personal digital assistants (PDAs), Tablet PCs, and mobile phones using a stylus \cite{kristensson2004shark2,zhai2003shorthand}, has become one of the dominant text entry methods on modern touchscreen devices. 
Its success is attributed to the ease of learning \cite{kristensson2007discrete,zhai2003shorthand,zhai2012word} and the high entry rates achievable \cite{kristensson2007discrete,kristensson2004shark2,reyal2015performance}. 

Leiva et al.~\cite{leiva2021we} conducted a study to collect word-gesture typing data from 909 users. As this study focused solely on data collection, no statistical decoding process was implemented in the keyboard. Expert users achieved a typing speed of 50 words per minute (WPM), while those unfamiliar with gesture keyboards reached 40 WPM. In contrast, Reyal et al.~\cite{reyal2015performance} conducted a separate study with 12 participants using Google Keyboard (Gboard) to assess text entry speeds. In this study, the entry rate for participants increased from 33.6 WPM in the initial block to 39.1 WPM by the ninth block. Despite the difference in participant numbers, with the latter study involving only 12 participants, a clear gap in text entry rate (50 WPM vs 40 WPM) is evident between the actual word-gesture keyboard text entry rates and the ideal rates assuming perfect decoding. 
This discrepancy underscores the critical need for the development of a more accurate decoder to enhance performance.

Given the benefits of word-gesture keyboards on mobile devices, researchers have explored adapting this input method for use in AR and VR environments~\cite{chen2019exploring,gupta2019rotoswype,shen2021simulating,wang2021investigating,yanagihara2019text,markussen2014vulture,yu2017tap,xu2019pointing}.
In these studies, the entry rates for the most proficient participants or expert users typically span from 20 to 40 WPM. 
A commonality among these methods is the utilization of the SHARK$^2$ decoder, chosen for its ease of configuration,
as these studies assert their novelty lies in the unique interaction modalities they introduce. 
However, the potential for enhanced performance through the adoption of a more advanced decoder is acknowledged, albeit with the caveat of requiring substantial implementation efforts. 
Therefore, there is a need for an advanced decoder ensuring both high decoding accuracy and plug-and-play capability in academic and research applications.

\subsection{Word-Gesture Decoders}
In this section, we first formally define the word-gesture decoding problem. 
Then, we introduce two commonly used approaches for word-gesture decoding: template-matching decoders and neural decoders. 
Lastly, we discuss the advantages and disadvantages of the two types of decoder, highlighting the need for an efficient joint approach.


\subsubsection{Problem Formulation}

When a cursor moves on a virtual keyboard, the movement is captured as a word-gesture trajectory, denoted by $g$. 
The initial step involves encoding this trajectory into a structured representation, $E(g)$.
One commonly used encoding approach involves using the Cartesian coordinate positions of the trajectory at sequential time intervals (illustrated in Figure~\ref{fig:trajcteory}), expressed as $E(g) = [x_1, x_2, ..., x_T],$ where $T$ is the length of the resampled cursor trajectory, and each $x_t \in \mathbb{R}^2$ is the location of the cursor at time $t$. 
Both the current template-matching decoder and neural decoder primarily use this encoding approach.

Following this, word-gesture decoding is mathematically articulated as finding $w^* = \arg\max_{w} P(w|E(g))$, which essentially means determining the word $w$ that, given the encoded trajectory $E(g)$, has the highest probability of being the intended input.

\subsubsection{Template-Matching Decoder}
Template-matching algorithm is a process that involves comparing an input trajectory $g$ with a set of pre-defined word-gesture templates \(P\) to identify the best match. 
\(P = \bigcup_{i=1}^{n} P_i\) is constructed from a lexicon \(C = \{w_1, w_2, ..., w_n\}\) where $n$ denotes the number of unique words in the lexicon.
Each word \(w_i\) in the lexicon is mapped to a word-gesture template \(P_i\) through a specific process (\(\text{ConstructTemplate}\)). 
A naive (\(\text{ConstructTemplate}\)) process is to sequentially connect the central points of each letter within a word \(w_i\) on a keyboard to generate a simple template \(P_i\).
More specifically, for a given encoded input trajectory $E(g)$, the algorithm computes a similarity score $S(E(g), P_i)$ (eg. Euclidean distance) for each pre-defined template $P_i$.
The algorithm then selects the template $P_j$ with the highest similarity score as the best match for the encoded trajectory $E(g)$: $P_j = \underset{i \in \{1,2,...,n\}}{\arg\max} \; S(E(g), P_i).$

SHARK$^2$~\cite{kristensson2004shark2} is one of the most frequently used template-matching decoders. The SHARK$^2$ algorithm employs a sophisticated template-matching technique that leverages a multi-channel architecture to enhance recognition accuracy for word-gestures~\cite{kristensson2004shark2}. Through these mechanisms, SHARK$^2$ efficiently narrows down the vast space of potential matches by quickly discarding non-viable candidates and focusing on those most likely to be correct. This multi-channel approach, combining shape and location information with a smart pruning strategy, allows SHARK$^2$ to offer high accuracy and speed in word-gesture decoding, even with a large vocabulary of possible inputs.

\subsubsection{Neural Decoder}
A neural decoder~\cite{hochreiter1997long,shen2023fast} is represented by a function \(D_{\theta}\), which maps an input \(y\) to a probability distribution over the alphabet for each time step. This function is parameterized by weights and biases \(\theta\), which are learned during training.
The model is trained using the Connectionist Temporal Classification (CTC) loss~\cite{graves2012connectionist}, which aligns the output sequence \(\pi\) with the target label sequence \(z\) by maximizing the probability of \(z\) given \(\pi\) across all possible alignments, considering insertions of the CTC blank label \(\phi\) where necessary.

For each time step \(t\), the model outputs a probability distribution \(\pi_t\) over the extended character alphabet \(L'_{\text{char}}\). The extended alphabet includes all the characters in the model's alphabet plus a special token for the CTC blank label, denoted as \(\phi\). 
If the original alphabet \(L_{\text{char}} = \{a, \ldots, z\}\) represents 26 lowercase English letters, then the extended alphabet \(L'_{\text{char}} = L_{\text{char}} \cup \{\phi\}\) includes these letters plus the blank token.
The output \(\pi_t\) for each time step \(t\) is a vector in the simplex \(\Delta^{|L'_{\text{char}}|}\), meaning that it represents a probability distribution across the extended alphabet. The dimension of \(\pi_t\) is \(|L'_{\text{char}}|\), where \(|L'_{\text{char}}|\) is the size of the extended alphabet. If we consider just the lowercase letters plus the blank token, \(|L'_{\text{char}}| = 27\).

The complete output of the model \(\pi\) for the entire swipe gesture is a sequence of these probability distributions across all time steps, so \(\pi = [\pi_1, \pi_2, \ldots, \pi_T]\).
Each \(\pi_t \in \Delta^{|L'_{\text{char}}|}\), making the dimension of \(\pi\) to be \(T \times |L'_{\text{char}}|\).
Figures~\ref{fig:conventional_decoder} and~\ref{fig:pixelated_decoder} both visualize the probability distributions \(\pi\) using heatmaps.

Formally, the neural decoder model can be mathematically represented as follows:

\[D_{\theta}: \mathbb{R}^{m \times T} \rightarrow (\Delta^{|L'_{\text{char}}|})^T\]

where \(D_{\theta}(y) = \pi\) and each \(\pi_t \in \Delta^{|L'_{\text{char}}|}\) for \(t = 1, \ldots, T\).

\subsubsection{Comparison Between SHARK$^2$ and Neural Decoder}
\label{sec:comparison}

\begin{s_itemize}

\item \textbf{SHARK$^2$ is Easier to Configure and Implement.}

SHARK$^2$ supports customizable (\(\text{ConstructTemplate}\)) process, thus it enables the adaptability to different technologies.
In contrast, a neural decoder requires a large dataset for effective training, especially for novel WGK systems in AR and VR.
One could theoretically utilize existing word-gesture data from Mobile Phone WGK~\cite{leiva2021we} to adapt the data to different keyboard sizes through various transformation functions. 
However, as illustrated in Figure ~\ref{fig:teaser}, the intrinsic properties of the data may significantly differ, resulting in suboptimal performance.
Additionally, creating a neural decoder demands deep learning expertise, and costly resources on complex training and hyperparameter tuning processes.

\item  \textbf{Neural Decoder has Significantly Better Decoding Accuracy.}

Despite SHARK$^2$'s ease of configuration, SHARK$^2$ suffers from persistent performance gaps compared to neural decoders.
Shen et al.~\cite{shen2023fast} demonstrated a substantial improvement in accuracy when employing a neural decoder, achieving a low error rate of 5.41\%, as opposed to the higher Character Error Rate (CER) of 35.34\% found with SHARK$^2$.
Character Error Rate is the percentage of characters that were incorrectly predicted compared to the total number of predicted characters.
Naturally, for efficient text entry, high prediction accuracy from word-gesture decoders is essential. 

\item  \textbf{ SHARK$^2$ is a Word-Level Model and Neural Decoder is Character-Level Model.}

SHARK$^2$ uses word-gesture templates from a lexicon, enabling predictions only for words within this set. 
Thus, SHARK$^2$ is a word-level model.
Conversely, a neural decoder, embodying a character-level model, predicts characters sequentially without relying on a set of pre-defined templates.
This character-level model can progressively predict with each character word-gesture, offering enhanced flexibility and immediacy in text entry.

\end{s_itemize}

\begin{figure}[t]
    \centering
    \includegraphics[width=1\textwidth]{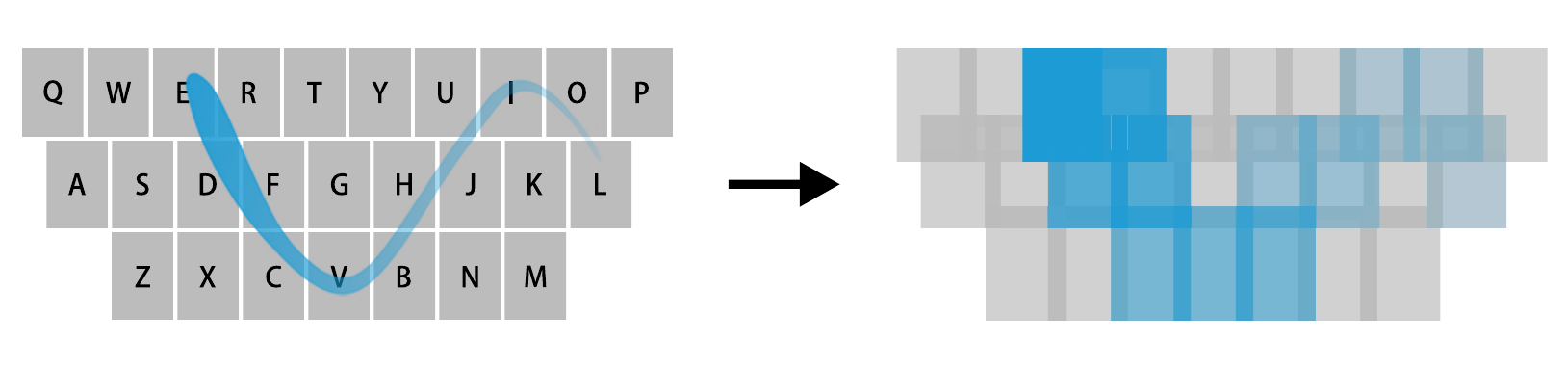}
    \caption{The discretization process converts a continuous word-gesture trajectory into discrete `pixels' via a mapping function. The `pixels' are larger than key tabs to compensate for ambiguous and noisy trajectories. 
    }
    \label{fig:discretization}
\end{figure}

\section{Pre-Training a Neural Decoder into a Generalizable Model}
As outlined in the comparison between SHARK$^2$ and the conventional neural decoder discussed in Section~\ref{sec:comparison}, there is a critical need for a model to achieve a balance between ease of configuration and high accuracy in word-gesture decoding.
To this end, we introduce a model that combines the simplicity of setup with robust performance, thereby mitigating the weaknesses of each of the former approaches.
Drawing inspiration from large language models known for their zero-shot learning capabilities, which are achieved through pre-training on vast corpora~\cite{brown2020language, kojima2022large}, we aim to pre-train a neural decoder with a substantial volume of data to achieve generalizability for different WGK systems. 
However, direct training using the Cartesian coordinate sequence $x$ utilized by both SHARK$^2$~\cite{kristensson2004shark2} and the neural decoder~\cite{alsharif2015long} is impractical, as these sequences are fine-grained and vary significantly across different WGK systems in AR and VR.

We propose a novel encoding method that encodes the continuous trajectory sequence into a coarse discretized representation, as illustrated by Figure~\ref{fig:discretization}. 
This approach addresses the challenge of dataset variability and enhances our model's understanding of trajectory patterns.
In the following sections, we will elaborate on our discretization methodology, provide details of our training data, and describe the specialized architecture of our model designed to optimize pre-training effectiveness.

\subsection{Word-Gesture Trajectory Discretization}
\label{sec:discretization}
Encoding is the process of transforming unstructured data into structured elements that a computer can process. 
In previous attempts to encode word-gesture trajectories~\cite{kristensson2004shark2,alsharif2015long, shen2023fast}, the strategy involved utilizing Cartesian coordinate positions.
From an information theory perspective, using coordinate positions directly for word-gesture decoding has limitations due to their continuous nature, whereas the desired output (characters) is discrete. 
Continuous data can vary infinitely within a given range, leading to a high degree of uncertainty and requiring more information to specify each point precisely. 
In contrast, discrete outputs, such as characters, have a finite set of possibilities. 
This mismatch means that directly mapping continuous input to discrete output can introduce inefficiencies and inaccuracies, necessitating algorithms to effectively bridge this gap by discretizing the continuous input or employing strategies to reduce the information loss during this conversion.

Our method discretizes a word-gesture trajectory into coarse discretized `pixel' regions on the keyboard, a process we refer to as word-gesture trajectory discretization.
Figure~\ref{fig:discretization} demonstrate this discretization process and Figure~\ref{fig:pixelated_traj} visualize the discretized trajectory in one-hot-encoding for the word `quickly' using a heatmap.
Instead of tracking continuous movement, we assign each segment of a word-gesture trajectory to the corresponding `pixel' region it traverses according to a mapping function. 
This approach simplifies complex word-gesture trajectories into a sequence of discrete `pixels.' 
This discretization enhances the accuracy of recognizing word-gesture patterns by minimizing the impact of noise in position tracking, variations in user behavior and keyboard sizes, thereby making the neural decoder more efficient at predicting user input. 
This method effectively bridges the gap between the continuous nature of word-gesture trajectory and the discrete structure of text input, enabling more accurate and efficient decoding of word-gesture trajectories. Formally:

\begin{s_enumerate}
    \item Discretizing the keyboard into discretized regions based on the positions of the keys, where each region is defined by a square with height and width proportional to the key tab's dimensions, adjusted by a ratio factor.
    \item Define a mapping function \(C(x)\) that maps a given trajectory point \((x_i)\) to a region based on the discretized regions it falls within. This function embodies the discretization process, assigning each point on the keyboard to a specific region.
    \item The discretized trajectory \(T\) can be expressed as \(T = \{C(x) \mid (x_i) \in S\}\), where each element \(C(x_i)\) is the region corresponding to the segment of the trajectory passing through that region on the keyboard. The discretized trajectory is then one-hot encoded and used as the input for the neural decoder. 
\end{s_enumerate}

This mathematical formulation encapsulates the process of transforming continuous word-gesture trajectory into discrete sequences of `pixels'.

\subsection{Training Dataset}

We pre-trained the pixellated neural decoder using two datasets: 
\begin{s_enumerate}
    \item \textit{Mobile Phone WGK Dataset}: 
    Our research utilizes a public dataset, `how we swipe'~\cite{leiva2021we}, gathered through a web-based custom virtual keyboard on mobile devices. This dataset includes 8,831,733 touch points corresponding to 11,318 unique English words gestured by 1,338 users, with 11,295 unique words correctly gestured and 3,767 words gestured inaccurately.
    \item \textit{Synthetic Dataset}: Our study integrates the GAN (Generative Adversarial Network)-Imitation model proposed by Shen et al.~\cite{shen2021imaginative}, which they subsequently applied this model to synthesize word-gesture trajectories~\cite{shen2021simulating}. 
    They then performed extensive evaluations and comparisons of the GAN-Imitation model with other techniques for generating synthetic word-gesture trajectory data to train neural decoders~\cite{shen2023fast}. 
    For our research, we adopted the synthetic strategy detailed in ~\cite{shen2023fast}, training the synthetic model using the \textit{Mobile Phone WGK Dataset}. 
    In this paper, we do not analyze synthetic data generation methods, instead concentrating our efforts on examining the discretization and pre-training approaches.
\end{s_enumerate}

We have constructed a large-scale training dataset comprising 95,649 trajectory samples from the \textit{Mobile Phone WGK Dataset}, alongside 100,000 trajectory samples from the \textit{Synthetic Dataset}. This dataset consists of 32,347 unique words.

\subsection{Implementation Details}

Here is a detailed description of our model and the associated training specifics. 
We conducted a comprehensive hyperparameter optimization process to determine the values of the hyperparameters~\cite{Bergstra2011AlgorithmsFH}.
\begin{s_itemize}
    \item \textbf{Model Configuration}:
    We use PyText~\cite{pytext} to implement our model. 
    The core of our model comprises a Bi-directional Long Short-Term Memory (BiLSTM)~\cite{sundermeyer2012lstm,hochreiter1997long} layer, which is crucial for understanding the temporal dependencies within the input sequences. 
    This representation layer consists of two stacked LSTM layers with a hidden dimension of 222, allowing the model to capture both forward and backward context effectively.
    To prevent overfitting, a dropout rate of 0.3 is applied within this layer, providing regularization by randomly omitting a subset of features during training.
    Following the creation of the representation layer, a dense fully connected layer serves as an intermediary, facilitating the transition from the LSTM output to the decoder. 
    This layer employs a Rectified Linear Unit (ReLU)~\cite{Agarap2018DeepLU} activation function, layer normalization~\cite{Ba2016LayerN}, and an additional dropout rate of 0.3 to maintain regularization.
    The final component of our model is the decoder layer, which utilizes a Connectionist Temporal Classification (CTC)~\cite{graves2012connectionist} beam decoder. 
    This decoder implements a beam search algorithm~\cite{Freitag2017BeamSS} to efficiently explore the most probable word candidates by evaluating combinations of character probabilities at each timestep. 
    Beam search enhances the model's ability to predict sequences accurately by considering multiple hypotheses concurrently.
    \item \textbf{Training Details}:
    For training our model, we employed the Adam~\cite{kingma2014adam} optimizer with a learning rate of 0.01, complemented by a minimal weight decay of 0.00001 to prevent overfitting further. The training process was conducted over 600 epochs, with an early stopping mechanism disabled to allow the model to fully converge. The batch size for training, evaluation, and testing was uniformly set to 128 to balance computational efficiency and training stability.
    \changetext{The model was initialized with random weights for training.}
    To accommodate the computational demands of training and ensure numerical stability, we adopted mixed-precision training using the `FP16OptimizerFairseq' from Fairseq~\cite{ott2019fairseq}, starting with an initial loss scale of 128. This approach not only accelerates training but also reduces the memory footprint, allowing for the use of larger batch sizes or models.
    Additionally, our training regimen included reporting metrics to TensorBoard~\cite{tensorflow2015-whitepaper} for real-time monitoring and analysis of the model's performance. 
    \changetext{
    The best model configuration, as determined by Top-4 word prediction accuracy, was automatically saved and loaded post-training to ensure that our results were based on the peak performance of the model. This configuration was achieved through a hyperparameter sweep, experimenting with over 100 configurations, including the number of LSTM layers, the dimension of the LSTM layer, the dropout rate, and the learning rate, to find the optimal model that achieves the highest accuracy while ensuring the model size remains below 5 MB.}


\end{s_itemize}

\begin{table*}[t]
\scriptsize
\centering
\begin{tblr}{
  cells = {c},
  hline{1-2,8} = {-}{},
}
                                & {Mobile Phone WGK} & {Mid-Air Poke (AR)}     & {Mid-Air Poke (VR)} & {Mid-Air Pinch (VR)} & {On-Surface WGK} \\
{Number of Users} &  1,338 & 16 &   200    &   200     &  100  \\
{Device Platform}  & \textit{Mobile Phone} & \textit{HoloLens 2}         &     \textit{Quest 2}  &    \textit{Quest 2}      &   \textit{PC}   \\
{Number of Unique Words}              & 11,295 & 1,700     &    3,000   &    3,000    &   2,123    \\
{Number of Samples}              & 95,649 & 25,513       &    51,413 & 51,433 & 33,594    \\    
{Text Entry Rates (WPM)}             & 31.11 (SD=12.49) & 21.39 (SD=13.17)     & 26.9 (SD=10.19)   &  27.4 (SD=9.88) &   24.7 (SD=11.14)  \\   
{Touch Points Statistics}              & 0.15 (0.27) &   0.042 (0.063)      &   0.030 (0.049)  & 0.028 (0.045) &  0.12 (0.23)   \\   
\end{tblr}
\caption{We gathered datasets from five unique WGK systems.
We report text entry rates with the mean and standard deviation (SD).
We report two touch point statistics: Average Distance to Key Center (ADKC) and Average Major Axis Length (AMAL) in the parentheses.
The computation of these two statistics is detailed in Section~\ref{sec:touch_point}.
}
\label{tab:datasets}
\end{table*}


\section{Validation Datasets}
We posit two hypotheses: firstly, that distinct WGK systems exhibit unique data patterns; and secondly, that our \textit{Pre-trained Neural Decoder} serves as a universally applicable solution across various WGK systems. 
To explore these hypotheses, we have collected an array of datasets from different WGK systems, encompassing a range of interactions including on-surface touch, mid-air poking (pointing), and mid-air pinching (raycasting)~\cite{Mifsud2022AugmentedRF} across both AR and VR platforms featuring keyboards of different sizes.
Table~\ref{tab:datasets} gives an overview of the four datasets. 

\subsection{Mid-Air WGK in VR}
\label{sec:vr_data}

\begin{figure}[t]
  \centering
  \begin{subfigure}[b]{\textwidth}
  \centering
    \includegraphics[width=0.5\textwidth]{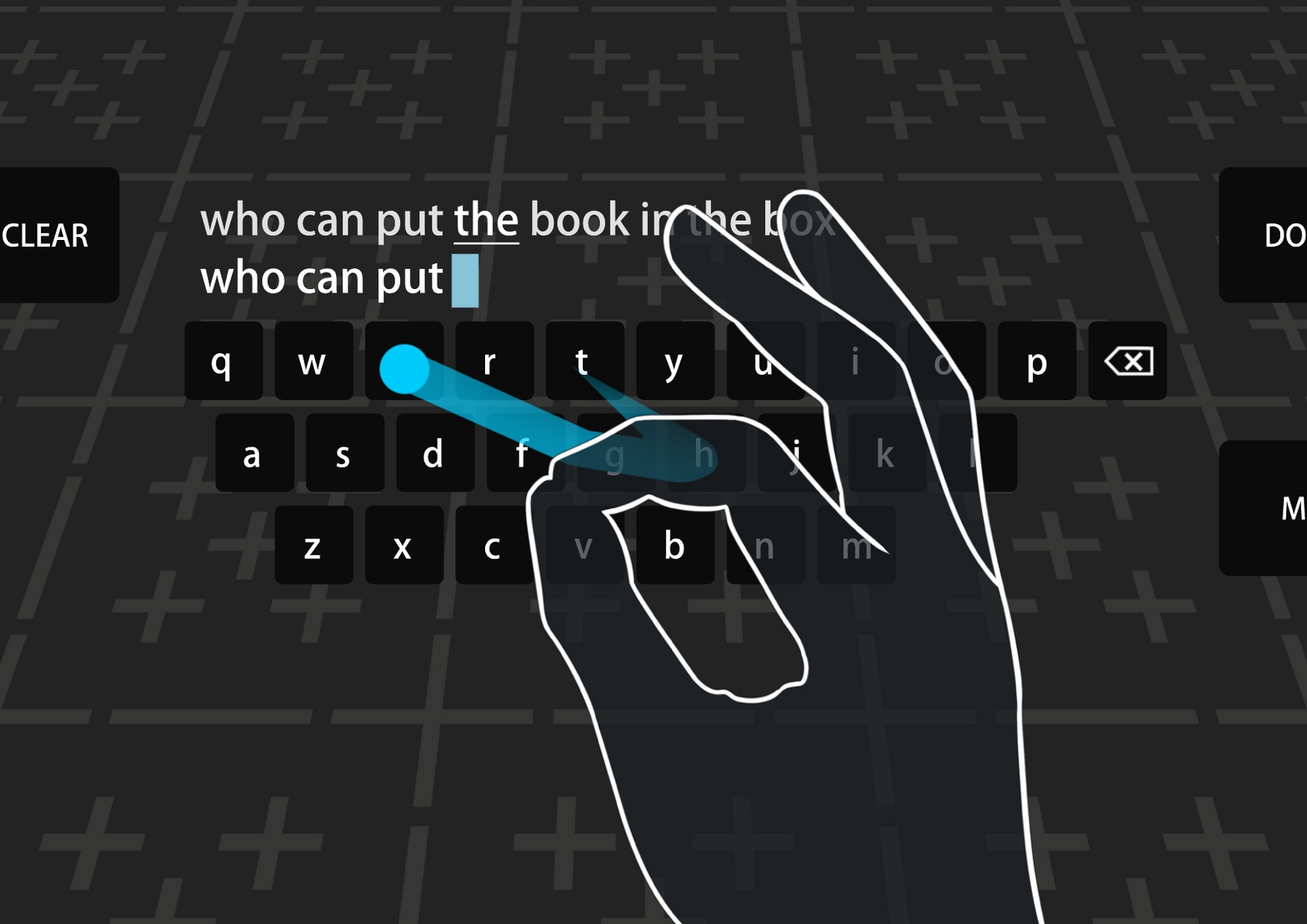}
    \caption{VR mid-air pinch WGK.}
    \label{fig:vr_pinch_study}
  \end{subfigure}
  \begin{subfigure}[b]{\textwidth}
  \centering
    \includegraphics[width=0.5\textwidth]{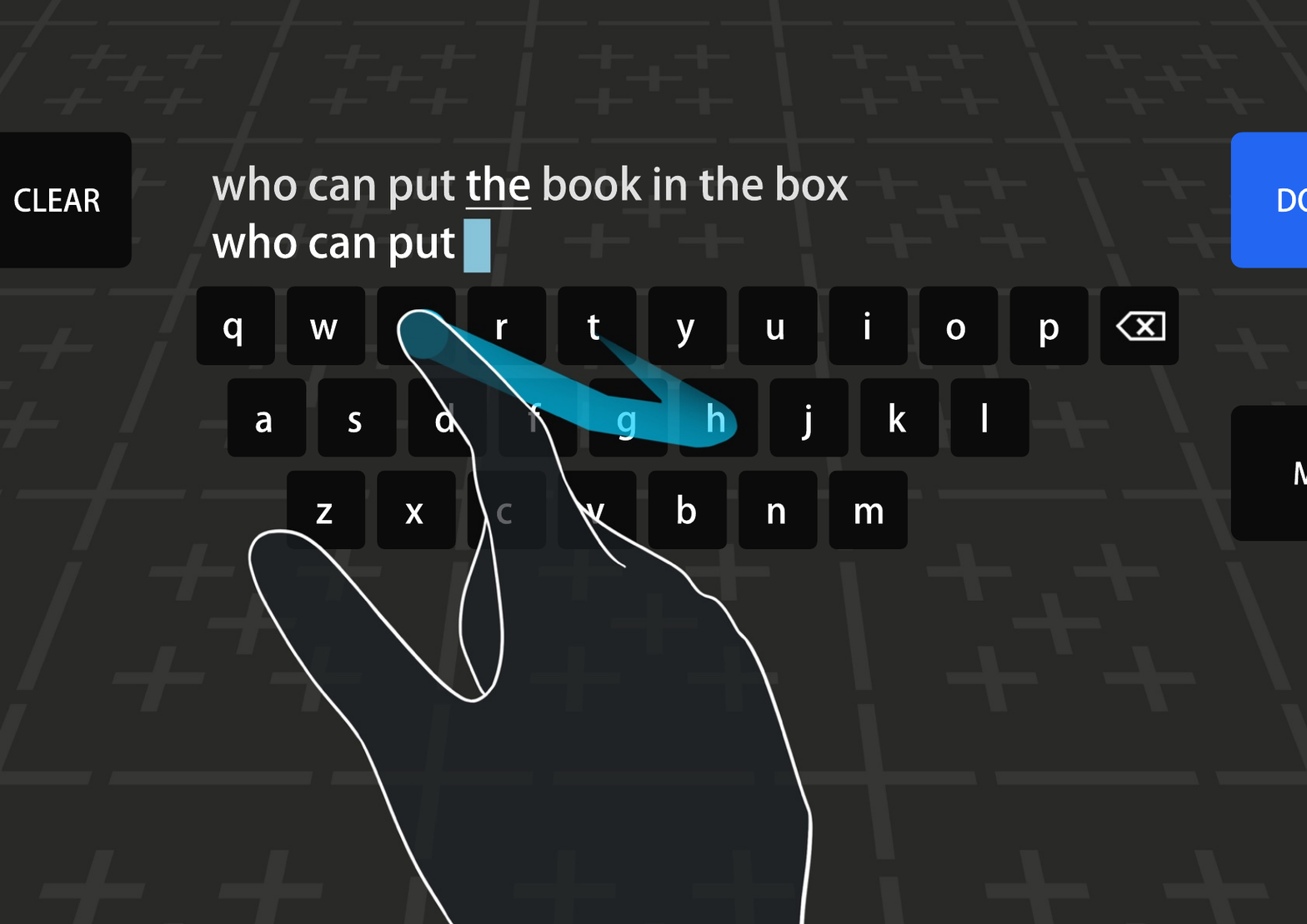}
    \caption{VR mid-air poke WGK.}
    \label{fig:vr_poke_study}
  \end{subfigure}
  \caption{Data collection of 200 participants for mid-air word-gesture keyboards (WGK) in VR. }
\end{figure}

We begin our exploration by examining WGKs within virtual reality (VR), a domain that has seen extensive investigation~\cite{dube2019text}.
Notably, platforms such as Quest Headsets~\cite{OculusQuestSeries} have integrated mid-air WGK functionalities in their v56 updates~\cite{Meta2023QuestV56}.
Our primary focus lies on two interaction modes: mid-air poke WGK and mid-air pinch WGK.
Despite extensive studies exploring these two interactions~\cite{dudley2018fast,speicher2018selection}, there are no public datasets available on VR mid-air WGK. 
To address this gap, we undertook our own data collection, involving a substantial scale of 200 users. The details of our study are as follows:

\begin{s_enumerate}
    \item \textbf{Participants}: 
    We recruited 200 volunteers as participants through an internal mailing list, with an age distribution as follows: 46 in their 20s, 74 in their 30s, 51 in their 40s, and 29 in their 50s. 
    The group comprised 100 males and 100 females. 
    We collected responses from participants about the frequency of their word-gesture typing usage on mobile phones: 20\% had rarely or never used it, 32\% sometimes used it (at least once a month), 26\% often used this feature (at least once a week), and 22\% always used the feature (at least once a day).
    \item \textbf{Procedure}: Data collection was conducted in the participants' homes using the retail Quest 2, with the Android Application Package (APK) delivered via the experiment app on the Oculus Store.
    Participants were randomly assigned to two groups, each experiencing two conditions: pinch first and poke first. 
    Before these conditions, users were introduced to the study through a step-by-step tutorial explaining the experiment.
    Figure~\ref{fig:vr_pinch_study} and Figure~\ref{fig:vr_poke_study} demonstrate the word-gesture typing technique with pinch and poke interaction modes used in the study. 
    The experiment application uploaded the study logs via a Dropbox API upon completion. 
    We employed a `Wizard of Oz' decoding strategy, akin to that described in Shen et al.~\cite{shen2022personalization}, to guarantee the integrity of the data collected. 
    Each condition for one participant takes around 30 minutes to complete.
    \item \textbf{Phrase Set}: The phrase dataset was curated from the GLUE~\cite{wang2018glue}, MacKenzie~\cite{mackenzie2003phrase}, and Enron~\cite{klimt2004enron} datasets, resulting in a collection of 2,700 unique phrases comprising 3,000 unique words.
    No other information except text entry data was collected during the study, and the text entry data was anonymized. 
\end{s_enumerate}

\subsection{Mid-Air WGK in AR}
\label{sec:mid_air_ar}
The distinction between AR and VR in the context of mid-air WGK lies in the interaction feedback when directly poking the virtual keyboard: AR allows for `real' hand interaction with a virtual keyboard, whereas VR facilitates interaction through a virtual hand with a virtual keyboard. 
Consequently, VR interaction is more precise because users can accurately see the virtual hand's position relative to the virtual keyboard, creating a closed-loop feedback system from the user's physical hand to the virtual hand control. 
Conversely, in AR, the absence of a virtual hand and the inaccuracies in hand tracking lead to an open-loop nature in hand interaction and control, potentially resulting in noisier input data~\cite{kim2023star}.
In contrast, AR and VR mid-air pinch word-gesture typing operate similarly, offering a closed-loop system of projected remote cursor control on the virtual keyboard without significant differences. 
Therefore, in AR, the focus is on poke-based mid-air WGK.

We utilized two public datasets regarding AR mid-air pinch word-gesture typing, which were taken from the studies conducted by Shen et al.~\cite{shen2022personalization, shen2023fast}.
In one study, Shen et al.~\cite{shen2022personalization} introduced AdaptiKeyboard, a personalizable mid-air WGK for AR specifically designed for HoloLens 2. 
This keyboard employs multi-objective Bayesian optimization to dynamically adjust the keyboard size, aiming to optimize both speed and accuracy concurrently.
The data collected for this study involved word-gesture typing on various sizes gathered from 12 participants.
In a separate study, Shen et al.~\cite{shen2023fast} presented a novel mid-air WGK design that removed visual feedback and relaxed the delimitation threshold.
Throughout their user studies, they collected word-gesture typing data using different interaction designs from 34 participants.

\subsection{On-Surface WGK }

\begin{figure}[t]
    \centering
    \begin{subfigure}[b]{\textwidth}
    \centering
    \includegraphics[width=0.5\textwidth]{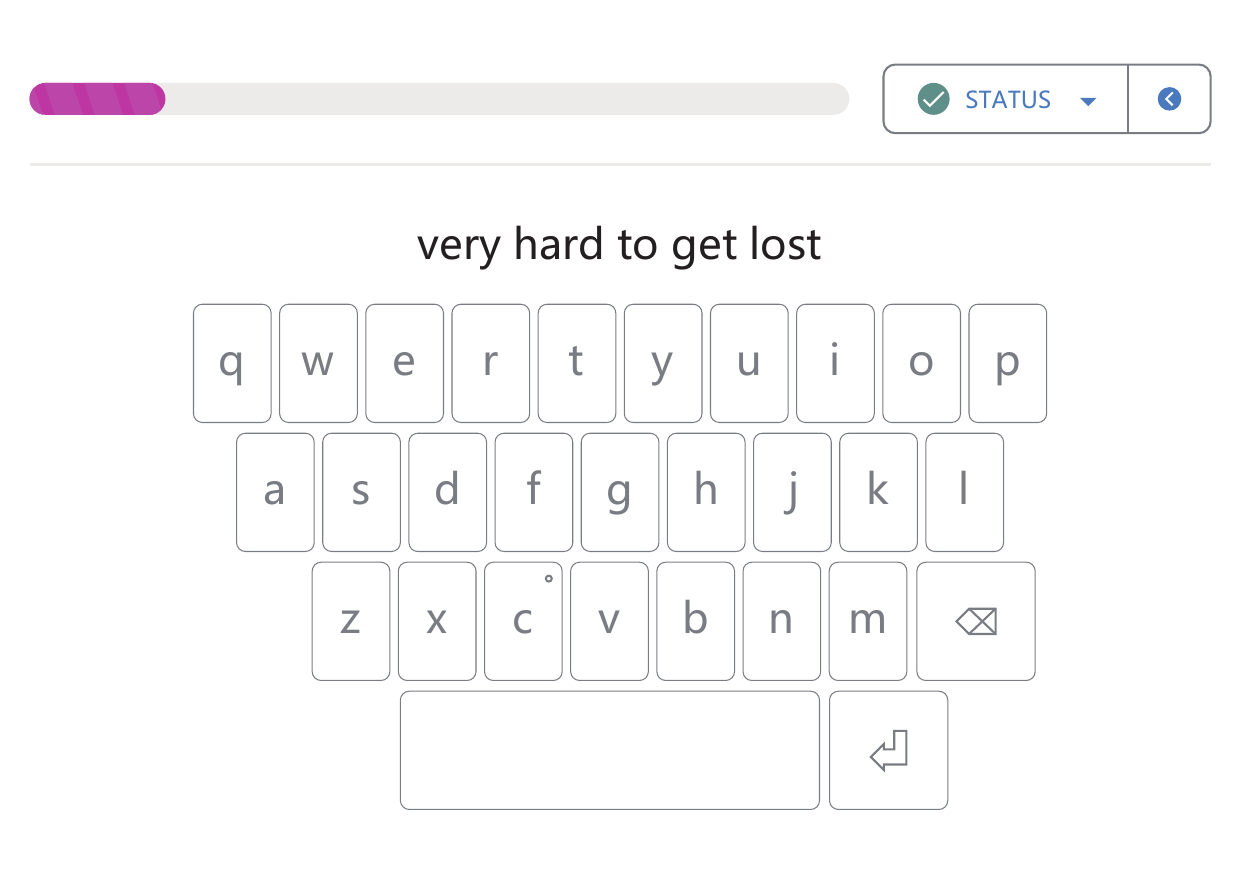}
    \subcaption{Study page showing device connectivity, study progression, and keyboard with the cursor.}
    \label{fig:study_frontend}
    \end{subfigure}
    \hfill 
    \begin{subfigure}[b]{\textwidth}
    \centering
    \includegraphics[width=0.5\textwidth]{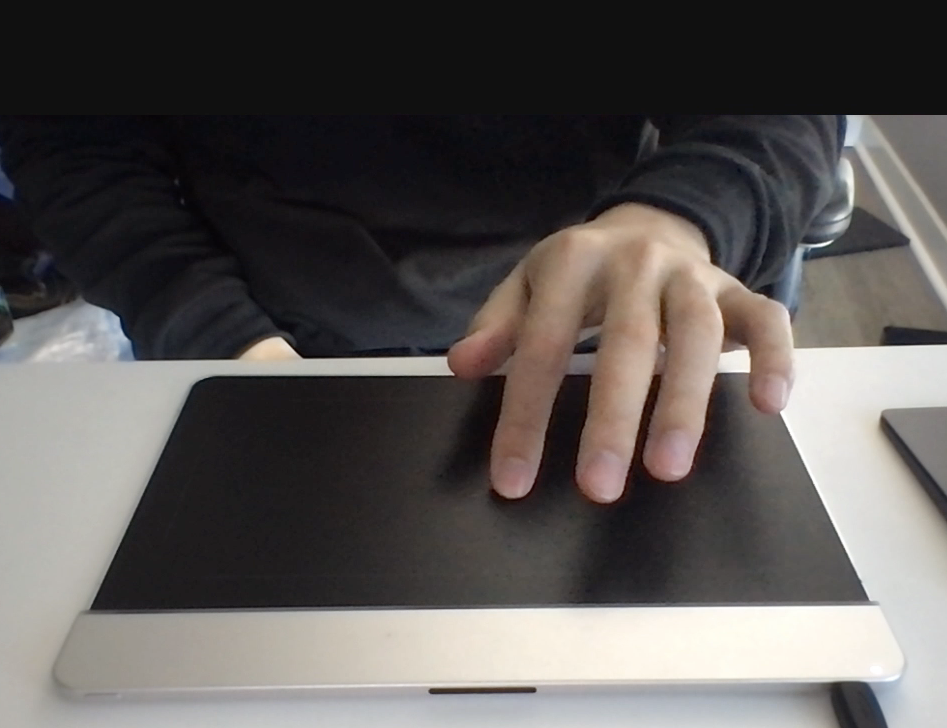}
    \subcaption{ We use a haptic touchpad with a capacitive grid~\cite{sensel_haptic} to collect ground truth trajectory data.
}
    \label{fig:sensel_trackpad}
    \end{subfigure}
    \caption{Data collection of 100 participants for on-surface word-gesture keyboards (WGK).}
\end{figure}

Apart from mid-air WGK, we also investigate on-surface WGK, which can be utilized in both AR and VR environments. 
On-surface WGK requires clear delimitation, similar to how WGK is performed on touchscreens; that is, placing a finger on the surface signals the beginning of a trajectory, and lifting it off the surface signals the end of the trajectory. 
To collect on-surface WGK data, we opt for alternatives to the built-in hand tracking from AR/VR headsets, as the current computer vision-based hand tracking cannot accurately detect the on-surface delimitation. 
Therefore, we use Sensel’s haptic touchpad with capacitive touch and force field sensors~\cite{sensel_haptic} to collect ground truth trajectory data (Figure \ref{fig:sensel_trackpad}). 
Moreover, instead of using a headset for display, we employ a monitor screen connected to a personal computer (PC), showing a virtual keyboard. 
This setup fully simulates a virtual touchpad, which can be on the palm~\cite{liang2015ar}, or any other surface in an AR environment~\cite{Apple2023UseMacWithVisionPro, Mifsud2022AugmentedRF}.
This data collection resulted in a total of 33,594 trajectories, encompassing 2,123 unique words from 100 participants.

The following provides further details on the data collection process:

\begin{s_enumerate}
    \item \textbf{Participants}: 
    We recruited 100 volunteers as participants through an internal mailing list, with an age distribution as follows: 19 in their 20s, 44 in their 30s, 24 in their 40s, and 13 in their 50s. 
    The group comprised 64 males, 33 females, and 3 participants who preferred not to disclose their gender. 
    We collected responses from participants about the frequency of their word-gesture typing usage on mobile phones: 12\% participants had rarely or never used it, 33\% sometimes used it (at least once a month), 43\% often used it (at least once a week), and 12\% always used the feature (at least once a day).
    \item \textbf{Procedure}: Participants were seated in front of computer screens that displayed a keyboard interface, complete with a cursor and its trace, as illustrated in Figure~\ref{fig:study_frontend}. They interacted with the interface using Sensel’s haptic touchpad (Figure~\ref{fig:sensel_trackpad}) to facilitate word-gesture typing of prompted words on the screen. Each participant was tasked with word-gesture typing 160 words, randomly selected from an extensive phrase set, to ensure broad data collection. 
    To mitigate fatigue, participants could take breaks of up to two minutes after completing sets of 10 words. We developed a front-end interface using Lab.js~\cite{henninger2022labjs}, which visualized the cursor and its trace and automatically generated random phrase prompts from the phrase set. Similarly to the method employed in Section~\ref{sec:vr_data}, we utilized a `wizard of oz' decoder for data collection.
    Each session for one participant takes around 1 hour to complete.
    \item \textbf{Phrase Set}: The phrase dataset was created from two distinct sources: the Enron Mobile Corpus~\cite{klimt2004enron} and the MacKenzie phrase set~\cite{mackenzie2003phrase}, resulting in a collection of 2,320 unique phrases comprising 2,123 unique words.
\end{s_enumerate}

\section{Word-Gesture Data Analysis}
We investigate the word-gesture typing datasets through two analytical perspectives: the assessment of touch point distributions and the evaluation of a geometric feature: curvature. 

\subsection{Touch Point Distribution Analysis}
\label{sec:touch_point}

\begin{table*}[t]
\scriptsize
\centering
\begin{tblr}{
  cells = {c},
  cell{1}{2} = {c=2}{},
  cell{1}{4} = {c=2}{},
  cell{1}{6} = {c=2}{},
  cell{1}{8} = {c=2}{},
  hline{1,6} = {-}{0.08em},
  hline{2-3} = {-}{},
}
                            & Mid-Air Poke (AR)                                                            &                                                                    & Mid-Air Poke (VR)                                                            &                                                                    & Mid-Air Pinch (VR)                                                           &                                                                    & On-Surface WGK                                                           &                                                                    \\
                            & Top-1                                                              & Top-4                                                              & Top-1                                                              & Top-4                                                              & Top-1                                                              & Top-4                                                              & Top-1                                                              & Top-4                                                              \\
SHARK$^2$~\cite{kristensson2004shark2}                   & {$43.2\% $\\($\pm 3.2\%$)} & {$51.3\% $\\($\pm 2.4\%$)} & {$50.3\% $\\($\pm 3.9\%$)} & {$55.9\% $\\($\pm 3.1\%$)} & {$48.9\% $\\($\pm 2.5\%$)} & {$54.7\% $\\($\pm 2.7\%$)} & {$42.1\% $\\($\pm 4.1\%$)} & {$49.2\% $\\($\pm 4.3\%$)} \\
Conventional Neural Decoder~\cite{alsharif2015long} & {$74.6\% $\\($\pm 1.8\%$)}   &  {$81.0\% $\\($\pm 1.3\%$)}                                          & {$79.8\% $\\($\pm 2.0\%$)}         &                                {$86.3\% $\\($\pm 1.9\%$)}        & {$77.9\% $\\($\pm 2.1\%$)}                                               &       {$84.8\% $\\($\pm 2.4\%$)}                           &        {$71.5\% $\\($\pm 2.2\%$)}                                                            &                   {$81.6\% $\\($\pm 1.9\%$)}                                                 \\
\textbf{\textit{Pre-trained Neural Decoder}}  & {$82.5\% $\\($\pm 1.4\%$)}     &            {$89.8\% $\\($\pm 1.1\%$)}                              & {$85.1\% $\\($\pm 1.9\%$)}         &                                {$91.9\% $\\($\pm 1.4\%$)}     & {$82.7\% $\\($\pm 1.5\%$)}                                              &       {$90.2\% $\\($\pm 1.0\%$)}                                &                {$83.0\% $\\($\pm 2.0\%$)}                                                    &         {$89.5\% $\\($\pm 1.8\%$)}                                                           
\end{tblr}
\caption{Decoding accuracy on four datasets of our proposed \textbf{\textit{Pre-trained Neural Decoder}} compared to two baselines: SHARK$^2$ and conventional neural decoder. Results are reported in Top-1 and Top-4 accuracy, standard deviation is reported in parentheses.}
\label{tab:decoding_accuracy}
\end{table*}

We utilized the visualization methods from Chen et al.~\cite{chen2023DRG} to compute the touch point distributions (visualized by plotting 95\% confidence ellipses for each key) across various validation datasets. 
Additionally, we compute the statistics of the touch point distributions for each dataset:
\begin{s_itemize}
    \item \textbf{Average Distance to Key Center (ADKC)}:
    The average distance from the center of each ellipse to the centers of the corresponding keys, again averaged over all keys. This metric assesses the bias in touch point locations relative to the intended target. The equation for computing this statistic is:
    \[
    \text{ADKC} = \frac{1}{N} \sum_{i=1}^{N} \sqrt{(x_{c,i} - x_{k,i})^2 + (y_{c,i} - y_{k,i})^2}
    \]
    where \((x_{c,i}, y_{c,i})\) denotes the center of the ellipse for key \(i\), and \((x_{k,i}, y_{k,i})\) represents the center of key \(i\).
    \item \textbf{Average Major Axis Length (AMAL)}:
    The average of the major axes lengths of the 95\% confidence ellipses, calculated across all keys. This quantifies the spatial dispersion of touch points for each key, providing a measure of touch accuracy and precision. The equation for this statistic is:
    \[
    \text{AMAL} = \frac{1}{N} \sum_{i=1}^{N} 2 \sqrt{\lambda_{i,max}}
    \]
    where \(\lambda_{i,max}\) represents the largest eigenvalue of the covariance matrix for the touch points on key \(i\), and \(N\) is the total number of keys.
\end{s_itemize}
The computation of these statistics is based on normalizing trajectory data to a unit-width keyboard for a fair comparison between keyboards of different sizes.
The plots for these distributions are displayed in the second row of Figure~\ref{fig:teaser}, and the statistics are presented in Table~\ref{tab:datasets}.
Our observations revealed distinct word-gesture typing interactions, each showcasing a unique distribution of touch points. 
Specifically, AR mid-air WGK exhibits more noise compared to VR WGK, attributable not only to the aforementioned distinction between AR and VR hand interactions in Section~\ref{sec:mid_air_ar}, but also to other device-dependent factors. 
First, the HoloLens 2 offers lower hand tracking accuracy than the Quest 2, affecting precision. Second, in optical see-through AR, users can see a virtual keyboard without occlusion of their physical hands, complicating the task of accurately locating the virtual keyboard in relation to their hands. 
This challenge is compounded by the fact that the HoloLens 2 does not perform hand segmentation, allowing users to see their hands both before and beyond the virtual keyboard simultaneously.
Lack of occlusion and the positioning of the virtual keyboard far from their hands introduce inaccuracies not only in tracking but also in user perception.

The `fat finger' problem becomes particularly noticeable with on-surface word-gestures and mobile phone word-gestures, marked by a scattered distribution of touch points and the larger AMAL and ADKC values. 
The term `fat finger' effect describes the difficulty users face when their fingers, which are relatively large compared to the keys or touchpoints, mistakenly press adjacent keys or register incorrect word-gestures.
This problem is a direct result of the small touch surfaces of mobile devices and touchpads.
Unlike mid-air keyboards, which are larger and thus more accommodating to the natural size of human fingers, the touch surfaces on mobile phones and touchpads are considerably smaller, making it challenging to hit the intended keys accurately.

\subsection{Geometric Feature Analysis}

\begin{figure}[t]
    \centering
    \includegraphics[width=0.6\textwidth]{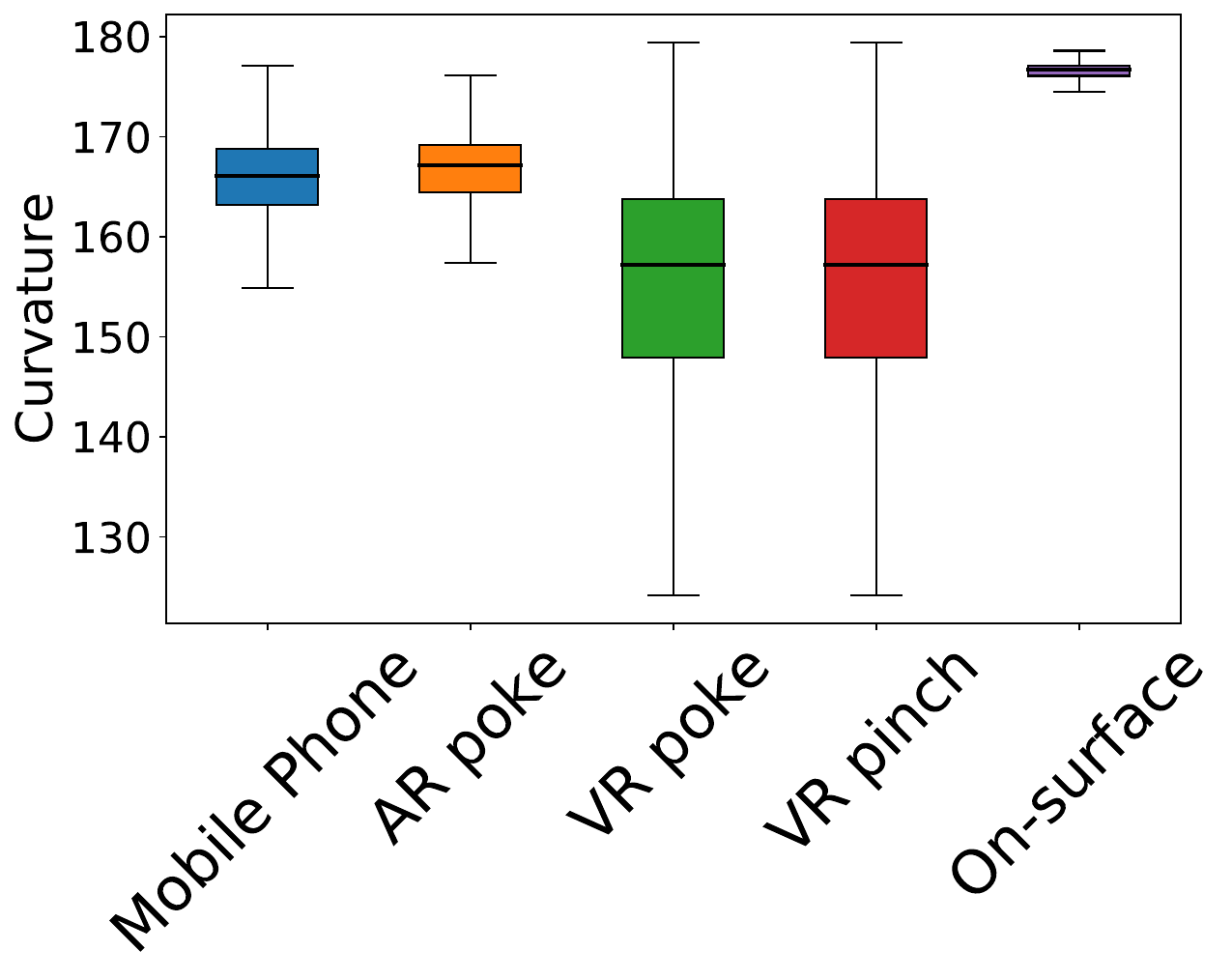}
    \caption{Box plots depicting mean, median, and quartiles of the curvature of the trajectories from the five datasets.}
    \label{fig:geometric_features}
\end{figure}  

We evaluate the geometric distinctions across datasets through a geometric feature: curvature~\cite{shen2021simulating}\footnote{Details of the computation of curvature can be found in~\cite{shen2021simulating}.}.

Curvature quantifies the local bending at any point along a curve, serving as an indicator of the degree to which it diverges from a straight line. 
A high curvature value signifies a pronounced bend, whereas a low curvature corresponds to a minor bend or a linear trajectory.

Figure~\ref{fig:geometric_features} showcases the diversity in curvature present across the datasets. 
We note that VR-based WGK exhibits a broader and lower curvature range, suggesting a predominance of more linear trajectories. 
This pattern could stem from extensive body movements in a large interaction space, which tend to produce more straightforward trajectories. 
Conversely, AR maintains less linear trajectories due to the challenges associated with the interaction between physical hands and virtual keyboard. 
Among the datasets analyzed, on-surface WGK displays the most pronounced curvature, indicative of highly curved trajectories. This observation is attributed to the less direct control over the cursor compared to the control mechanisms found in Mobile Phone WGK and Mid-Air WGKs.

\section{Experiments}
Our research uses a series of experiments to assess the efficacy of a \textit{Pre-trained Neural Decoder} across the four datasets. 
Initially, we evaluate the performance of the pre-trained decoder against benchmarks, specifically SHARK$^2$ and a conventional neural decoder, to establish a comparative baseline. 
Subsequently, we discuss the potential enhancements achievable through fine-tuning the decoder on specific datasets.
Furthermore, we conduct studies to examine the influence of different discretization techniques. 
Additionally, we explore alterations to the model's architecture to identify which structural configurations yield the optimal results for pre-training. 
Lastly, we assess the latency of the \textit{Pre-trained Neural Decoder}.

\subsection{Baselines Comparison}

\subsubsection{Baselines}
In this study, we evaluate the performance of a pre-trained decoder by comparing it with two established baselines: 
\begin{s_enumerate}
    \item SHARK$^2$  Decoder: For our evaluation, we adopt the parameter settings detailed by Kristensson et al.~\cite{kristensson2004shark2}, ensuring a direct comparison under standardized conditions.
    \item Conventional Neural Decoder: The backbone of the conventional neural decoder is the same as our \textit{Pre-Trained Neural Decoder}, with the only difference being the dimension of the input layer to accommodate the dimension of the Cartesian trajectory input.
\end{s_enumerate}
Following the setup, both baseline decoders and the \textit{Pre-trained Neural Decoder} are subjected to thorough testing on the designated test datasets. 
We employ the Leave-One-Subject-Out methodology for our experiments.
This structured approach allows for a comprehensive assessment of each decoder's capabilities in handling real-world data.

\subsubsection{Evaluation Measure}
We use Top-k accuracy to evaluate the models. Top-k accuracy measures the model's ability to predict the correct word within its Top-k predictions.
Formally, let $y_i$ be the true label for the $i$-th instance and let $P_{i, k}$ be the set of top $k$ predictions for the $i$-th instance made by the model. The Top-k accuracy over $N$ instances is defined as: $ \frac{1}{N} \sum_{i=1}^{N} \mathbf{1}\left(y_i \in P_{i, k}\right)$, where $\mathbf{1}(\cdot)$ is the indicator function, which is 1 if the condition is true and 0 otherwise. In our analysis, we specifically report on Top-1 and Top-4 accuracy, providing insights into both the accuracy of the model's primary prediction and its ability to offer relevant alternatives within the top four suggestions.


\subsubsection{Results}

Table~\ref{tab:decoding_accuracy} reveals that our \textit{Pre-trained Neural Decoder} significantly surpasses both the SHARK$^2$ decoder and the conventional neural decoder. 
The \textit{Pre-trained Neural Decoder} achieves an average Top-1 accuracy of 83.3\% and an average Top-4 accuracy of 90.4\%. 
Switching from the SHARK$^2$ decoder to a \textit{Pre-trained Neural Decoder} results in an average increase of 37.2\% for Top-1 accuracy and 37.6\% for Top-4 accuracy. Moving from a naively trained neural decoder to a \textit{Pre-trained Neural Decoder} leads to additional improvements in average accuracy, with a 7.4\% increase for Top-1 and a 6.9\% increase for Top-4.
The average accuracy of the \textit{Pre-trained Neural Decoder} across various tasks is 83.3\%. This Top-1 accuracy is already sufficient for research use. By equipping the \textit{Pre-trained Neural Decoder} with a bigram language model and an auto-correction module of vocabulary size over 50,000 ~\cite{Vertanen2019MiningAA, shen2023fast}, the average improvement is 11.2\%, resulting in a final average Top-1 accuracy of 94.5\%, which is an adequate accuracy rate for large-scale commercial use.

\subsection{Fine Tuning Improves Model Performance, but Marginally}

We conducted a comprehensive analysis to evaluate the impact of model fine-tuning. Our hypothesis suggests that fine-tuning, particularly when tailored to specific datasets, could markedly enhance the neural decoder's performance.
Each dataset typically exhibits unique features or distributions inherent to its domain. Fine-tuning enables the neural decoder to adjust its pre-existing representations to accurately capture these domain-specific attributes.
To validate our hypothesis, we rigorously fine-tuned the \textit{Pre-trained Neural Decoder} with the training set of each dataset and assessed the neural decoders using the respective test sets. 
Table~\ref{tab:fine_tune_decoding} demonstrates that fine-tuning effectively leverages the comprehensive learning acquired by the \textit{Pre-trained Neural Decoder}, channeling their extensive capabilities to address the unique characteristics presented by a new dataset.
The average improvement from the \textit{Pre-trained Neural Decoder} to the fine-tuned decoder across all datasets is 3.4\%.
However, the average improved performance is not as large as the difference between the \textit{Pre-trained Neural Decoder} to the conventional neural decoder and the SHARK$^2$. Therefore, the \textit{Pre-trained Neural Decoder} already achieves adequate accuracy for research purposes and becomes suitable for commercial use when equipped with a language model. Thus, fine-tuning is not necessary.

\begin{table}
\scriptsize
\centering
\begin{tblr}{
  cells = {c},
  hline{1-2,4} = {-}{},
}
                                & {Mid-Air \\ Poke (AR)}     & {Mid-Air \\ Poke (VR)} & {Mid-Air \\ Pinch (VR)} & {On-Surface \\ WGK} \\
{Pre-trained} & {$82.5\% $\\($\pm 1.4\%$) } &   {$85.1\% $\\($\pm 1.9\%$) }    &    {$82.7\% $\\($\pm 1.5\%$) }       &  {$83.0\% $\\($\pm 2.0\%$) }         \\
{Fine-tuned}  &   {$85.8\% $\\($\pm 1.2\%$) }          &     {$88.0\% $\\($\pm 1.5\%$) }     &    {$86.7\% $\\($\pm 1.8\%$) }       &   {$86.4\% $\\($\pm 2.3\%$) }         
\end{tblr}
\caption{Decoding accuracy after fine-tuning the \textit{Pre-trained Neural Decoder}.}
\label{tab:fine_tune_decoding}
\end{table}

\subsection{Word-Gesture Trajectory Discretization Analysis}
Word-gesture trajectory discretization consists of two main components: the mapping function and the index encoding method. 
We explore various mapping functions and index encoding methods independently.

\subsubsection{Mapping Function}

\begin{table}
\scriptsize
\centering
\begin{tblr}{
  cells = {c},
  hline{1-2,4} = {-}{},
}
                                & {Mid-Air \\ Poke (AR)}     & {Mid-Air \\ Poke (VR)} & {Mid-Air \\ Pinch (VR)} & {On-Surface \\ WGK} \\
{Square \\ Region} & {$82.5\% $\\($\pm 1.4\%$) } &   {$85.1\% $\\($\pm 1.9\%$) }    &    {$82.7\% $\\($\pm 1.5\%$) }       &  {$83.0\% $\\($\pm 2.0\%$) }         \\
{Ellipse \\ Region}  &   {$81.9\% $\\($\pm 1.5\%$) }          &     {$84.6\% $\\($\pm 1.9\%$) }     &    {$81.2\% $\\($\pm 1.9\%$) }       &   {$83.2\% $\\($\pm 1.7\%$) }         
\end{tblr}
\caption{Decoding accuracy of using different mapping functions to map segments of a word-gesture trajectory to discrete `pixel' regions.}
\label{tab:discretization_decoding}
\end{table}

To refine the concept of defining the mapping function \(C(x, y)\) which are the regions for the trajectory coarse discretization in virtual keyboards, we examine the utility of square and elliptical regions. 
\begin{s_itemize}
\item Square Regions: The condition for a point \((x, y)\) to be within a square region centered at \((x_c, y_c)\) with dimensions \(2w\) and \(2h\) is:$  |x - x_c| \leq w \text{ and } |y - y_c| \leq h$, where \(w\) and \(h\) are half the width and height of the rectangle, respectively. 
\item Ellipse Regions: For an elliptical region around a key, centered at \((x_c, y_c)\) with semi-major axis \(a\) and semi-minor axis \(b\), a point \((x, y)\) falls within this region if: $  \frac{(x - x_c)^2}{a^2} + \frac{(y - y_c)^2}{b^2} \leq 1$.
\end{s_itemize}
By incorporating these shapes into \(C(x, y)\), we adjust the function to account for the region's shape associated with each keyboard character. This involves first identifying the shape and parameters for each key's region, then applying the corresponding condition to map \((x, y)\) to its character. 
For the Square Region, we define \(w\) and \(h\) as twice the key width and height, respectively. Similarly, in the Ellipse Region, we also define the semi-major axis \(a\) and semi-minor axis \(b\) as twice the key width and height.
This decision is informed by the Mobile Phone WGK touch point distribution analysis.
The analysis indicates that the average ratio, derived from comparing the mean values of the semi-minor and semi-major axes of the 95\% Confidence Ellipses to the mean dimensions (width and height) of keys, is approximately 2.
Having larger `pixels' also makes the discretized trajectory more tolerant to spatial noise and improves the neural decoder's ability to handle ambiguity.

Table~\ref{tab:discretization_decoding} presents the results of decoding accuracy from different mapping functions. 
We observed that Square Regions perform slightly better on average than Ellipse Regions, but not significantly. Therefore, the decision to use Square Regions is based on computational expense. 
Computing the condition for being within/outside Square Regions is less computationally expensive compared to Ellipse Regions. 
This is because, for a square region centered at \((x_c, y_c)\) with side length \(2s\), checking if a point \((x, y)\) is inside is simple: \(|x - x_c| \leq s\) and \(|y - y_c| \leq s\). 
This involves basic arithmetic and logical operations, making it computationally light.
Conversely, determining if a point lies within an elliptical region centered at \((x_c, y_c)\) with semi-major axis \(a\) and semi-minor axis \(b\) requires a more complex formula: \(\frac{(x - x_c)^2}{a^2} + \frac{(y - y_c)^2}{b^2} \leq 1\). 
This involves squaring differences, dividing by the axes' squares, and summing the fractions, which are more computationally demanding tasks than those for square regions.

\subsubsection{Index Encoding}

\begin{table}[t]
\scriptsize
\centering
\begin{tblr}{
  cells = {c},
  hline{1-2,4} = {-}{},
}
                                & {Mid-Air \\ Poke (AR)}     & {Mid-Air \\ Poke (VR)}  & {Mid-Air \\ Pinch (VR)} & {On-Surface \\ WGK}\\
{One-Hot \\ Encoding} & {$82.5\% $\\($\pm 1.4\%$) } &   {$85.1\% $\\($\pm 1.9\%$) }    &    {$82.7\% $\\($\pm 1.5\%$) }       &  {$83.0\% $\\($\pm 2.0\%$) }         \\
{Integer \\ Encoding}  &   {$73.2\% $\\($\pm 2.7\%$) }          &     {$73.2\% $\\($\pm 3.2\%$) }     &    {$68.0\% $\\($\pm 2.6\%$) }       &   {$64.2\% $\\($\pm 3.0\%$) }         
\end{tblr}
\caption{Decoding accuracy of different index encoding approaches.}
\label{tab:indexing}
\end{table}

Word-gesture trajectory, even when discretized into pixels, still require representation through numerical values for machine learning models to effectively learn from them. This essential step is known as encoding.

In our exploration, we delve into various `pixel' index encoding methods: 
1. \textbf{Integer Encoding}: This method assigns a unique integer value to each `pixel'. 
For instance, `a' is represented by the number 1, `B' by the number 2, and so on. 
This simple yet effective technique ensures that each element is distinctly identifiable by a specific numerical value. 
2. \textbf{One-Hot Encoding}: One-hot encoding takes a different approach, where each `pixel' is represented by a vector. 
This vector contains all zeros except for a single one at the position corresponding to the pixel.
For a 26-letter alphabet, the letter `a' would be represented by a vector starting with [1, 0, 0, ..., 0], and `b' would follow as [0, 1, 0, ..., 0]. 
This method provides a clear, binary representation of each character, distinguishing each one within a high-dimensional space.

Table~\ref{tab:indexing} demonstrates that one-hot encoding performs better than integer encoding. 
Moreover, one-hot encoding avoids implying a numerical relationship between pixels. 
Integer encoding could lead the model to assume an ordinal relationship where none exists, potentially skewing the learning process.
One-hot encoding represents each `pixel' as a distinct, equally distant vector, facilitating more accurate predictions.

\subsection{Model Structure}
We conducted tests using a transformer~\cite{vaswani2017attention} as the representation layer to learn latent representations, contrasting it with an LSTM-based approach. 
We experimented with various transformer model hyperparameters, and our findings indicate that the transformer layer struggles to effectively learn the patterns, consistently achieving a Top-1 accuracy of below 60\%, rendering it impractical for use. 
This underperformance of transformers compared to LSTMs in word-gesture typing decoding may stem from their inferior capability to manage the precise, localized context inherent in word-gesture trajectories. 
The inherent sequence processing ability of LSTMs better captures temporal dependencies crucial for this task. 
Additionally, transformers may require even larger datasets with millions of training samples to avoid overfitting and their complex architecture might not suit the latency requirements of a real-time word-gesture decoder.

\subsection{Latency Analysis}
Despite their high performance, deep-learning models can sometimes be too large and incapable of running in real-time. This is critical for applications like word-gesture typing decoders, which must operate in real-time. To address the real-time performance of our model, we used TorchScript~\cite{paszke2019pytorch} to script and export the model. 
Additionally, we applied PyTorch's quantization tool to reduce the model size significantly without sacrificing the accuracy~\cite{paszke2019pytorch}.
\changetext{This is possible because we report Top-K and Top-1 word prediction accuracy, evaluating the model's predictions at the word level, even though the model predicts a sequence of characters. After quantization, the model's performance increased by an average of 0.8\% in character error rate (CER), which measures the percentage of incorrectly predicted characters. However, when measured in Top-1 word prediction accuracy, which is calculated as \(100 - \text{word error rate (WER)}\), the accuracy remained unchanged because evaluating at the word level has a coarser granularity than evaluating at the character level.}
The quantized model is only 4 MB in size and can run with a latency of 97 ms on a Quest 3.
Generally, latencies below 100 ms are considered good for real-time interactions because they are perceived as nearly instantaneous by users.

\begin{figure*}[t]
\centering
\begin{subfigure}[b]{0.24\textwidth}
  \centering
  \includegraphics[width=\textwidth]{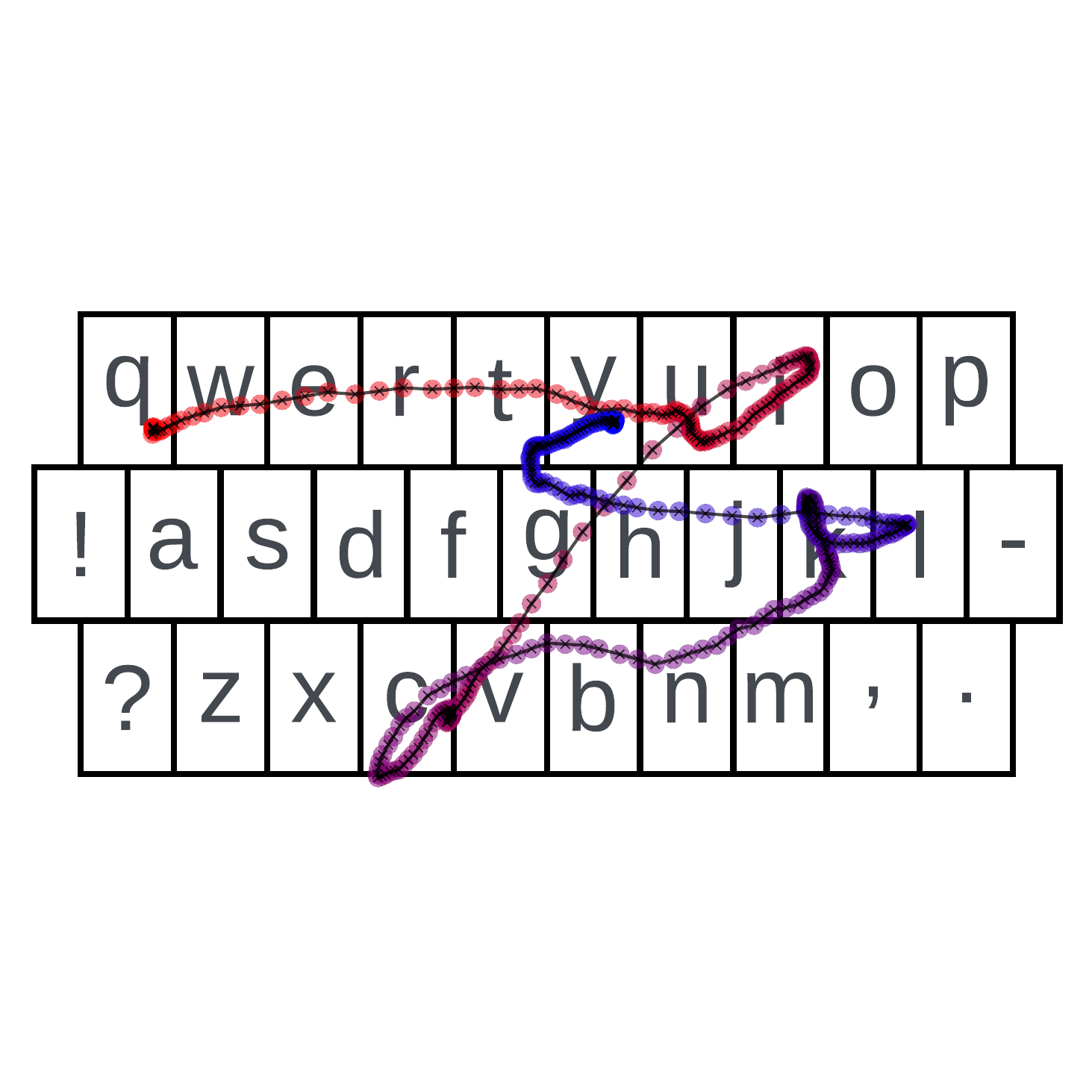}
  \subcaption{Cartesian Trajectory for `quickly' on QWERTY Keyboard.}\label{fig:trajcteory}
\end{subfigure}
\hfill
\begin{subfigure}[b]{0.24\textwidth}
  \centering
    \includegraphics[width=\textwidth]{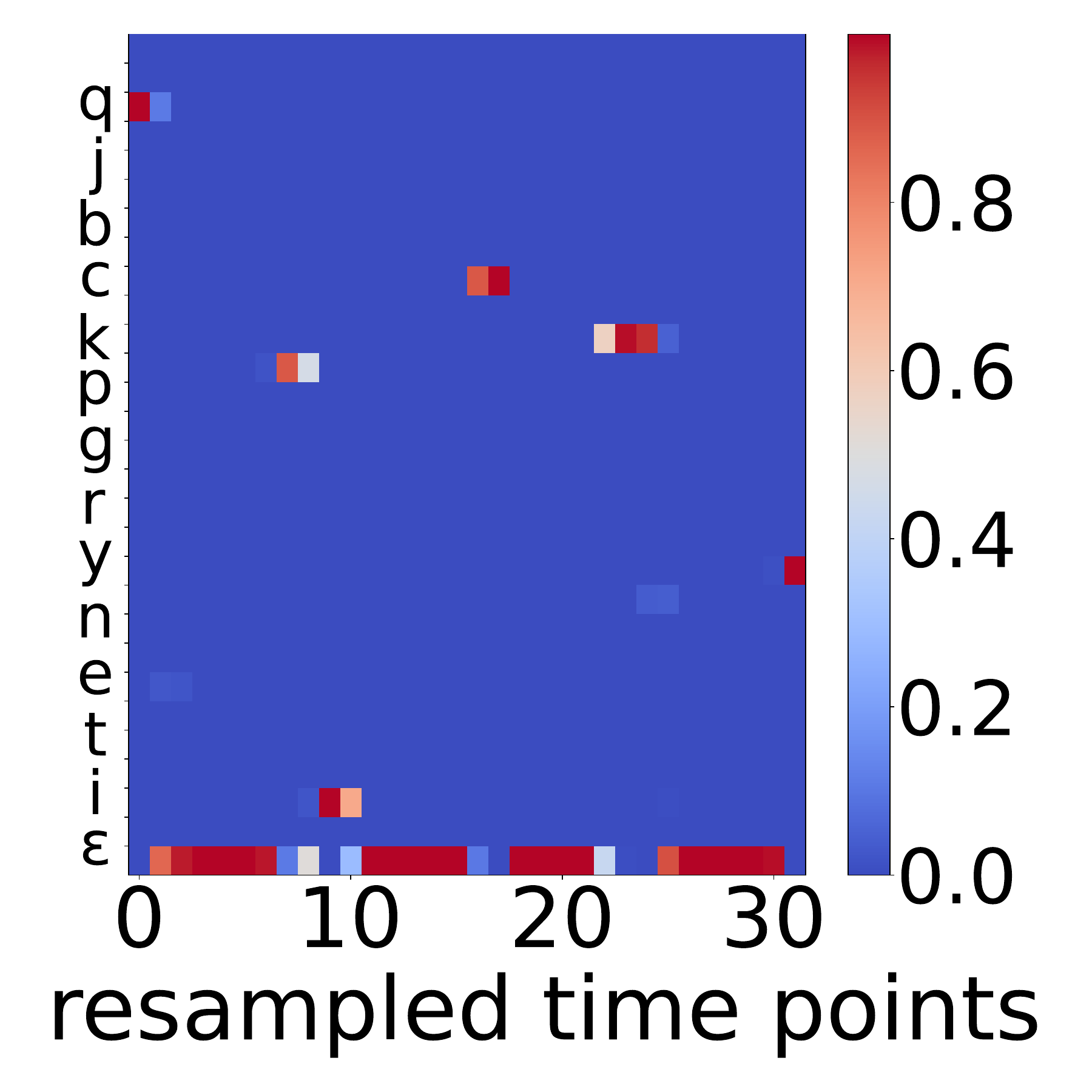}
  \subcaption{Heatmap Visualizing Model Output from the Conventional Neural Decoder.}\label{fig:conventional_decoder}
\end{subfigure}
\hfill
\begin{subfigure}[b]{0.24\textwidth}
  \centering
  \includegraphics[width=\textwidth]{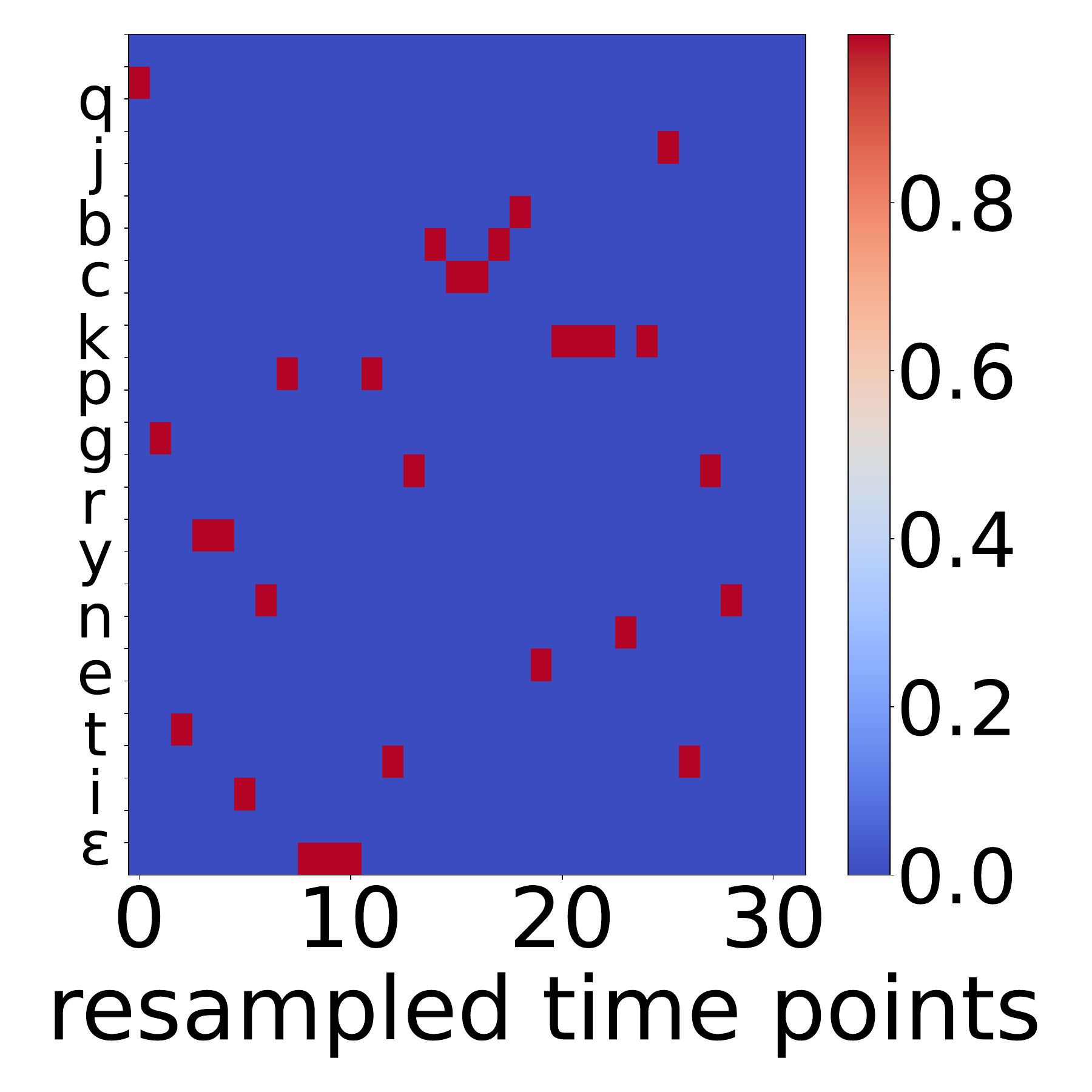}
  \subcaption{Heatmap Visualizing One-Hot Encoding of the Coarsely Discretized Trajectory for `quickly'.}\label{fig:pixelated_traj}
\end{subfigure}
\hfill
\begin{subfigure}[b]{0.24\textwidth}
  \centering
  \includegraphics[width=\textwidth]{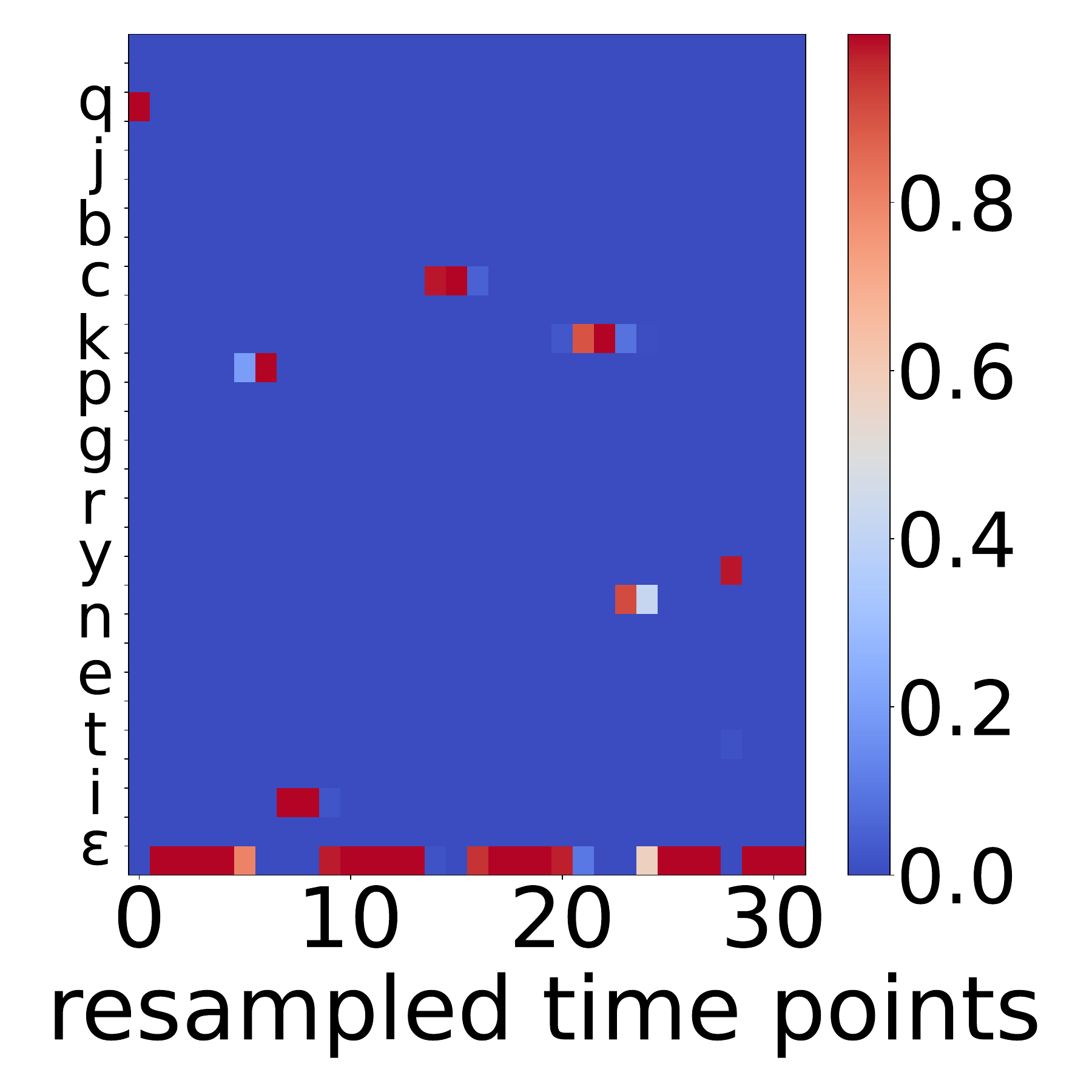}
  \subcaption{Heatmap Visualizing Model Output from the \textit{Pre-Trained Neural Decoder}.}\label{fig:pixelated_decoder}
\end{subfigure}
\caption{
This illustration provides a visual comparison of both the input \(y\) and output \(\pi\) for the conventional neural decoder and \textit{Pre-Trained Neural Decoder}, specifically in the context of decoding the word `quickly'.
}
\label{fig:illustration_decoding}
\end{figure*}

\section{Discussion}
In this section, we discuss the following different aspects: 
\begin{s_itemize}
    \item 
    \textbf{Contribution and Novelty}:
    In this study, we propose a novel approach to pre-training a neural decoder for word-gesture typing that demonstrates remarkable generalizability across various systems. By employing a unique discretization method to encode word-gesture trajectories, our model effectively learns from a vast dataset of real and synthetic data. This pre-training enables the decoder to accurately predict user input across diverse word-gesture typing systems in AR and VR environments without the need for fine-tuning. The significance of our contribution lies in developing a universal solution that combines the ease of configuration with high decoding accuracy, addressing the limitations of existing approaches such as SHARK$^2$ and conventional neural decoders.
    \item 
    \textbf{Why Our Method Works}:
    The success of our method can be attributed to the discretization of word-gesture trajectories. By converting continuous trajectories into discrete `pixels', we simplify the input space and allow the model to focus on learning the essential patterns of user input, as illustrated by Figure~\ref{fig:illustration_decoding}. 
    \changetext{Notably, the conventional neural decoder predicts only `quick', whereas the \textit{Pre-Trained Neural Decoder} accurately predicts `quickly'.
    This is illustrated in Figures~\ref{fig:conventional_decoder} and ~\ref{fig:pixelated_decoder}, where red 'pixels' signify high confidence in the character on the y-axis at the time step t on the x-axis. 
    This distinction is highlighted through the visualization of inputs: the Cartesian trajectory (Figure~\ref{fig:trajcteory}) and the one-hot encoding of the coarsely discretized trajectory (Figure~\ref{fig:pixelated_traj}). 
    The one-hot encoding aligns with the neural decoder's output space, sharing the same dimensions in terms of the number of classes and time length, which facilitates learning.
    In contrast, for the conventional neural decoder, the input (i.e., the Cartesian trajectory) exists in a continuous space, differing from the output's discrete space representation, thereby complicating the learning process.}
    \item 
    \textbf{Adoption to Touch-Type Decoding}:
    While our study primarily focuses on word-gesture typing, the proposed approach can easily adapt to touch-type decoding. The discretization method can be applied to individual key presses, representing each touch point as a discrete `pixel.' The model can learn to predict the intended characters based on the spatial distribution of touch points by training the neural decoder on a large dataset of touch-typing data. This adaptation would enable developing a robust and accurate touch-type decoder that can handle the challenges posed by the `fat finger' problem and variations in user typing patterns. 
    \item \textbf{Discretization as a Technique in other Applications}:
    In addition to its application in word-gesture typing, the discretization of continuous input has the potential to benefit various other machine learning problems. By converting continuous input into discrete representations, such as `pixels' or bins, the complexity of the data pattern can be reduced, making it more manageable for machine learning algorithms. Potential applications that can benefit from the discretization of continuous input include eye tracking, where discretization can simplify the interpretation of eye movement data, and affective computing, where discrete emotional states can be mapped from continuous physiological signals.
    \item \changetext{\textbf{The Potential Impact of Head-Mounted Display (HMD) Hand Tracking Accuracy on Word Prediction}: 
    Current video see-through HMDs, such as Meta Quest~\cite{OculusQuestSeries} and Apple Vision Pro~\cite{AppleVisionPro}, offer robust hand tracking with real-time visual feedback, showing rendered illustrations of hands for mid-air interactions. For example, a rendered hand appears when poking a virtual keyboard, or a cursor shows during a mid-air pinch, creating a closed-loop interaction. This visual feedback allows users to adjust their hand movements, reducing the impact of inaccurate tracking on text entry accuracy. In contrast, optical see-through HMDs like HoloLens~\cite{hololens} enable open-loop interaction by relying on natural hand perception without rendered visuals. Shen et al.~\cite{shen2023fast} investigated this by removing the visual feedback of a projected cursor on a mid-air gesture keyboard in HoloLens 2, finding that word prediction accuracy remained unaffected due to the robustness of conventional neural decoders. Our proposed \textit{Pre-trained Neural Decoder}, tested on Shen et al.~\cite{shen2023fast}'s dataset without visual feedback, achieved a Top-4 test accuracy of 90.1\%, demonstrating its ability to handle hand tracking inaccuracies in HMDs effectively.}

\end{s_itemize}

\section{Limitations and Future Work}

Our research, while comprehensive, is not without its constraints. 
The pre-trained decoder is optimized for QWERTY keyboards and may not perform as well with keyboards with different key arrangements.
Researchers actively explored new layouts such as the Metropolis~\cite{zhai2000metropolis}, Opti~\cite{MacKenzie1999Design} and Dvorak~\cite{Dvorak1936} for potential benefits in ergonomics, typing efficiency, or language accommodation, despite that these alternative layouts introduce learnability and adoption challenges for users~\cite{David1985Clio}. Our \textit{Pre-trained Neural Decoder} may not work seamlessly with alternative layouts out of the box, as the discretization process assumes a specific key arrangement with QWERTY sequential order. 

Furthermore, although we have successfully validated our model across four distinct datasets from various word-gesture typing systems in AR and VR environments, numerous other less common system designs exist, such as curved keyboards and eye-gaze-based word-gesture typing. While these were not explicitly tested, our model's functionality is rooted in its ability to process noisy and ambiguous trajectories.

Future work includes training the neural decoder on a larger, combined dataset and developing a transformer-based neural decoder. Initial tests with the transformer architecture showed poor performance, likely due to inadequate training data leading to overfitting. 
We aim to train the transformer-based decoder on a larger dataset and customize it to better accommodate long temporal patterns and meet latency requirements.

\section{Conclusion}

This paper introduces a novel \textit{Pre-trained Neural Decoder} that demonstrates remarkable versatility and accuracy for word-gesture typing across diverse AR/VR systems. By discretizing complex gesture trajectories into coarse `pixels' and pre-training on a large dataset, our model learns to accurately predict words from different datasets across various interaction modes and device platforms without requiring specific fine-tuning.
Extensive evaluations on four challenging datasets showcase the \textit{Pre-trained Neural Decoder}'s strong performance and generalizability, with an average Top-1 accuracy of 83.3\% and Top-4 accuracy of 90.4\% , which is a significant improvement over both conventional neural decoders and the popular SHARK$^2$ algorithm. The decoder also runs in real-time on mobile AR/VR hardware, enabling fluid gesture typing experiences. The proposed methodology illuminates a promising path towards universal \textbf{Gesture2Text} decoding that can enable efficient and expressive communication across new interactive contexts.

\bibliographystyle{abbrv-doi-hyperref}

\bibliography{template}

\appendix 

\section{About Appendices}
Refer to \cref{sec:appendices_inst} for instructions regarding appendices.

\section{Troubleshooting}
\label{appendix:troubleshooting}

\subsection{ifpdf error}

If you receive compilation errors along the lines of \texttt{Package ifpdf Error: Name clash, \textbackslash ifpdf is already defined} then please add a new line \verb|\let\ifpdf\relax| right after the \verb|\documentclass[journal]{vgtc}| call.
Note that your error is due to packages you use that define \verb|\ifpdf| which is obsolete (the result is that \verb|\ifpdf| is defined twice); these packages should be changed to use \verb|ifpdf| package instead.

\subsection{\texttt{pdfendlink} error}

Occasionally (for some \LaTeX\ distributions) this hyper-linked bib\TeX\ style may lead to \textbf{compilation errors} (\texttt{pdfendlink ended up in different nesting level ...}) if a reference entry is broken across two pages (due to a bug in \verb|hyperref|).
In this case, make sure you have the latest version of the \verb|hyperref| package (i.e.\ update your \LaTeX\ installation/packages) or, alternatively, revert back to \verb|\bibliographystyle{abbrv-doi}| (at the expense of removing hyperlinks from the bibliography) and try \verb|\bibliographystyle{abbrv-doi-hyperref}| again after some more editing.

\end{document}